\documentclass[letterpaper]{article} % DO NOT CHANGE THIS
\usepackage{aaai2026}  % DO NOT CHANGE THIS
\usepackage{times}  % DO NOT CHANGE THIS
\usepackage{helvet}  % DO NOT CHANGE THIS
\usepackage{courier}  % DO NOT CHANGE THIS
\usepackage[hyphens]{url}  % DO NOT CHANGE THIS
\usepackage{graphicx} % DO NOT CHANGE THIS
\usepackage{natbib}  % DO NOT CHANGE THIS AND DO NOT ADD ANY OPTIONS TO IT
\usepackage{caption} % DO NOT CHANGE THIS AND DO NOT ADD ANY OPTIONS TO IT
\usepackage{algorithm}

\usepackage{newfloat}
\usepackage{listings}
\DeclareCaptionStyle{ruled}{labelfont=normalfont,labelsep=colon,strut=off} % DO NOT CHANGE THIS
\floatstyle{ruled}
\newfloat{listing}{tb}{lst}{}
\floatname{listing}{Listing}
\title{VipAct: Visual-Perception Enhancement via Specialized VLM Agent Collaboration
and Tool-use}
\author{
    Zhehao Zhang\textsuperscript{\rm 1}\thanks{Work done during Zhehao's internship at Adobe.},  Ryan A. Rossi\textsuperscript{\rm 2}, Tong Yu\textsuperscript{\rm 2}, Franck Dernoncourt\textsuperscript{\rm 2}, Ruiyi Zhang\textsuperscript{\rm 2}, \\ Jiuxiang Gu\textsuperscript{\rm 2}, Sungchul Kim\textsuperscript{\rm 2}, Xiang Chen\textsuperscript{\rm 2}, Zichao Wang\textsuperscript{\rm 2}, Nedim Lipka\textsuperscript{\rm 2}
}
\affiliations{
    \textsuperscript{\rm 1}The Ohio State University\\
    \textsuperscript{\rm 2}Adobe Inc.
    zhang.16420@osu.edu,\\ \{ryrossi, tyu, dernonco, ruizhang, jigu, sukim, xiangchen, jackwa, lipka\}@adobe.com
}

\usepackage{bibentry}
\usepackage{microtype}

\usepackage{inconsolata}

\usepackage{graphicx}
\usepackage{algpseudocode}

\usepackage{url}

\usepackage{verbatim}
\newcommand{\cmark}{\textcolor{green!50!black}{\ding{51}}}
\newcommand{\xmark}{\textcolor{red!50!black}{\ding{55}}}
\usepackage{booktabs}
\usepackage{float}
\usepackage{colortbl}
\usepackage{xcolor}
\definecolor{lightred}{RGB}{255,102,102}
\definecolor{lightgreen}{RGB}{144,238,144}
\definecolor{Apricot}{rgb}{1.0, 0.85, 0.6}
\definecolor{BrickRed}{rgb}{0.8, 0.25, 0.33}
\definecolor{darkred}{rgb}{0.6, 0, 0}
\definecolor{darkgreen}{rgb}{0, 0.5, 0}
\usepackage[most]{tcolorbox} 
\usepackage{subcaption}
\usepackage{graphicx}
\usepackage{multirow}
\usepackage{caption}
\usepackage{svg}
\usepackage{nicefrac}
\usepackage{xcolor}
\usepackage{colortbl}
\usepackage{framed} 
\usepackage{soul}
\usepackage{nicefrac}
\usepackage{rotate}
\usepackage{adjustbox}
\usepackage{array}
\usepackage{capt-of}
\usepackage{tabulary}
\usepackage{amssymb}
\usepackage{amsmath}
\DeclareMathOperator*{\argmax}{arg\,max}
\usepackage{mathtools}
\usepackage{pifont} 
\usepackage{algorithm}
\usepackage{bm}
\usepackage{enumitem}
\usepackage{lipsum}

\usepackage{newfloat}
\usepackage{listings}
\DeclareCaptionStyle{ruled}{labelfont=normalfont,labelsep=colon,strut=off} % DO NOT CHANGE THIS
\floatstyle{ruled}
\newfloat{listing}{tb}{lst}{}
\floatname{listing}{Listing}
\begin{document}

\maketitle

 \begin{abstract}
While vision-language models (VLMs) have demonstrated remarkable performance across various tasks combining textual and visual information, they continue to struggle with fine-grained visual perception tasks that require detailed pixel-level analysis. Effectively eliciting comprehensive reasoning from VLMs on such intricate visual elements remains an open challenge. In this paper, we present \textsc{VipAct}, an agent framework that enhances VLMs by integrating multi-agent collaboration and vision expert models, enabling more precise visual understanding and comprehensive reasoning. \textsc{VipAct} consists of an orchestrator agent, which manages task requirement analysis, planning, and coordination, along with specialized agents that handle specific tasks such as image captioning and vision expert models that provide high-precision perceptual information. This multi-agent approach allows VLMs to better perform fine-grained visual perception tasks by synergizing planning, reasoning, and tool use. We evaluate \textsc{VipAct} on benchmarks featuring a diverse set of visual perception tasks, with experimental results demonstrating significant performance improvements over state-of-the-art baselines across all tasks. Furthermore, comprehensive ablation studies reveal the critical role of multi-agent collaboration in eliciting more detailed System-2 reasoning and highlight the importance of image input for task planning. Additionally, our error analysis identifies patterns of VLMs' inherent limitations in visual perception, providing insights into potential future improvements. \textsc{VipAct} offers a flexible and extensible framework, paving the way for more advanced visual perception systems across various real-world applications.
    \end{abstract}


    \section{Introduction}

Recent advances in large multimodal models (LMMs), particularly vision-language models (VLMs) \citep{gpt4o, bai2023qwenvl, chen2024internvl}, have shown impressive performance in integrating textual and visual information. Models like GPT-4o \citep{gpt4o} achieve strong results on various image-text benchmarks \citep{hudson2019gqa, lu2023mathvista, yue2024mmmu} and hold promise for real-world applications such as web navigation \citep{zheng2024gpt, he-etal-2024-webvoyager}. However, despite these advancements, studies \citep{rahmanzadehgervi2024vision, fu2024Blink, Tong_2024_CVPR, li2024densefusion} show that SOTA VLMs still struggle with fine-grained visual perception tasks, such as detecting line intersections or object boundaries—tasks that are trivial for humans. Overcoming these challenges is essential for deploying VLMs in critical applications like surgical robotics and autonomous driving, which demand precise visual understanding.

To address these challenges, prior works have explored visual programming methods \citep{subramanian-etal-2023-modular, Hu_2024_CVPR, gupta2023visual, suris2023vipergpt, mialon2023augmented, wu2023visual, yang2023mm}, where text queries are input into LLMs to generate code that invokes vision-specific models, using their outputs directly as predictions. While effective for predefined tasks, these methods lack generalizability beyond existing toolsets, limiting their use as universal visual perception solutions. Another line of research focuses on prompting strategies to elicit foundation models' System-2 reasoning by involving iterative reasoning with intermediate tokens \citep{yu2024distilling, saha2024system}. Textual prompting methods \citep{wei2022chain, saha2023branch, yao2024tree, besta2024graph} elicit LLMs to generate structured reasoning steps for complex text-based tasks, but their efficacy on fine-grained visual perception is underexplored. Similarly, visual prompting techniques \citep{lei2024scaffolding, yang2023set, wu2024dettoolchain}, which add artifacts like bounding boxes or masks to images, guide VLMs in interpreting visual data. While promising for some compositional visual reasoning, it is still unclear whether VLMs can accurately perceive such visual prompts, let alone whether these methods improve performance in visual perception.

To fill this gap, and inspired by advances in LLM-based agents \citep{wang2024survey, liu2023dynamic, AutoGPT2024, wang2024mixture, shen2024hugginggpt}, we propose \textsc{VipAct} (\textbf{VI}sual-\textbf{P}erception via VLM \textbf{A}gent \textbf{C}ollaboration and \textbf{T}ool-use), a VLM-based framework that integrates multi-agent collaboration and vision expert models for fine-grained visual perception tasks. As shown in Figure \ref{fig: framework}, \textsc{VipAct} consists of three core components: (1) an \textbf{orchestrator agent} that manages the workflow by analyzing tasks, coordinating agents, selecting tools, summarizing evidence, and deducing final answers; (2) \textbf{specialized agents} for tasks such as image captioning, visual prompt description, and image comparison, providing detailed visual analysis to the orchestrator; and (3) \textbf{vision expert models}, offering task-specific, fine-grained perceptual information to address VLMs' limitations. We evaluate \textsc{VipAct} against SOTA baselines across benchmarks that include diverse visual perception tasks featuring complex elements like visual prompts and multi-image inputs. \textsc{VipAct} consistently outperforms previous baselines on all tasks with different VLMs. Besides, our in-depth analysis highlights the importance of multi-agent collaboration in eliciting more detailed System-2 reasoning, as well as the critical role of visual input for task planning, with improved error handling and evidence aggregation.

Our key contributions are as follows: (1) \textsc{VipAct}, a multi-modal agent framework that synergizes multi-agent collaboration with vision expert models to enhance fine-grained visual perception. It is an autonomous system capable of handling diverse visual perception tasks using a single prompt template. It leverages a VLM for task analysis, planning, and invoking multi-agent collaboration, with flexible plug-and-play modular components. (2) We conduct experiments across diverse visual perception benchmarks, demonstrating \textsc{VipAct}'s advantages over SOTA baselines; (3) We systematically analyze previous methods that proved to be effective in improving the general capabilities of foundation models for fine-grained visual perception, revealing their inconsistent effectiveness. (4) We present comprehensive ablation studies to assess the impact of multi-agent collaboration, visual input for planning, and each component of \textsc{VipAct}, along with a detailed error analysis identifying the limitations of current VLMs, which serve as bottlenecks for further improvement.

\begin{table*}[!h]
 \centering
 % Use a smaller font size to help the table fit
 \small
 % Reduce column spacing to make room for all columns
 \renewcommand\tabcolsep{4.5pt} 
 \begin{tabular}{p{5.65cm}cccccccc}
 \toprule
 Methods & Reas. & Tool & Multi-Ag. & Plan Img & Exec Img & Img Loop & Multi-Img & Vis. Prompt \\
 \midrule
 ReAct \citep{yao2023react} & \cmark & \cmark & \xmark & \xmark & \xmark & \xmark & \xmark & \xmark\\
 MM-ReAct \citep{yang2023mm} & \cmark & \cmark & \xmark & \xmark & \cmark & \xmark & \xmark & \xmark\\
  ViperGPT \citep{suris2023vipergpt} & \xmark & \cmark & \xmark & \xmark & \cmark & \xmark & \xmark & \xmark\\
  VisProg \citep{gupta2023visual} & \xmark & \cmark & \xmark & \xmark & \cmark & \xmark & \xmark & \xmark\\
  CodeVQA \citep{subramanian-etal-2023-modular} & \xmark & \cmark & \xmark & \xmark & \cmark & \xmark & \xmark & \xmark\\
 \textbf{\textsc{VipAct} (Ours)} & \cmark & \cmark & \cmark & \cmark & \cmark & \cmark & \cmark & \cmark\\
 \bottomrule
 \end{tabular}
\caption{Comparison of \textsc{VipAct} with other LLM/VLM-based agentic frameworks. \cmark~ indicates the presence of a specific feature in the corresponding framework, \xmark~ its absence. Column abbreviations: ``Reas.'' for modules to elicit reasoning process, ``Tool.'' for tool integration, ``Multi-Ag.'' for multi-agent support, ``Plan Img'' for image input in planning, ``Exec Img'' for image input in execution, ``Img Loop'' for image use in iterative loops, ``Multi-Img'' for multi-image support, and ``Vis. Prompt'' for specific design for images containing visual prompts.}
\label{tab:difference}
\end{table*}

\section{Related Work} \label{sec:related-work}
\medskip\noindent\textbf{VLM-based Agent.} Advancements in LLM capabilities like planning have spurred the development of LLM-based agents across diverse applications \citep{zhang-etal-2023-crt, xi2023rise, chen2023autoagents, AutoGPT2024, shen2024hugginggpt, deng2024mind2web, zhang-etal-2024-e5, xie2024travelplanner, liu2023dynamic, liu2023agentbench, zhang2023exploring, zhou2023webarena}. The introduction of visually capable models has positioned them as backbones for vision-centric agents \citep{hu2024visual2}. Current research largely focuses on Web/GUI agents for interface interaction \citep{yan2023gpt, yang2023appagent, zheng2024gpt, xie2024osworld, kapoor2024omniact, zhang2024ufo, koh2024visualwebarena, wang2024mobile, lu2024weblinx, zhang2024android, deng2024multi, you2024ferret, zheng2024agentstudio, fan2024read, wang2024mobile2, he2024webvoyager} and embodied agents controlling robots \citep{nasiriany2024pivot, tan2024towards, ma2024survey, xie2024large, Yang_2024_CVPR, szot2024grounding}. However, VLM-based agents specifically for natural image perception tasks remain unexplored.

\medskip\noindent\textbf{Visual Programming.} Recent LLMs excel at code generation \citep{gao2023pal, zhang-etal-2023-crt, zhang-etal-2024-e5, zhang2024darg, schick2024toolformer}, enabling them to solve reasoning tasks via tool use, especially in areas like mathematical reasoning \citep{cobbe2021gsm8k, hendrycks2021measuring}. This paradigm has been extended to vision tasks \citep{subramanian-etal-2023-modular, Hu_2024_CVPR, gupta2023visual, suris2023vipergpt, mialon2023augmented, wu2023visual, koo-etal-2024-proptest}. Systems like MM-REACT \citep{yang2023mm} integrate LLMs with vision experts following ReAct's \citep{yao2023react} prompt template, while ViperGPT \citep{suris2023vipergpt} and VisProg \citep{gupta2023visual} use LLMs to generate executable code for visual reasoning without extra training. However, these often depend solely on text queries for code generation and employ rigid tool selection, hindering adaptation to new tasks. This limits their application to simpler visual QA scenarios \citep{hudson2019gqa, suhr-etal-2019-corpus, marino2019ok}, lacking support for fine-grained perception, visual prompts, or multi-image inputs, thereby restricting their utility in more complex visual reasoning. Table \ref{tab:difference} provides a detailed comparison.

\section{VipAct Framework} \label{sec: method}

\begin{figure*}[h!]
\centering
\includegraphics[width=1.0\linewidth]{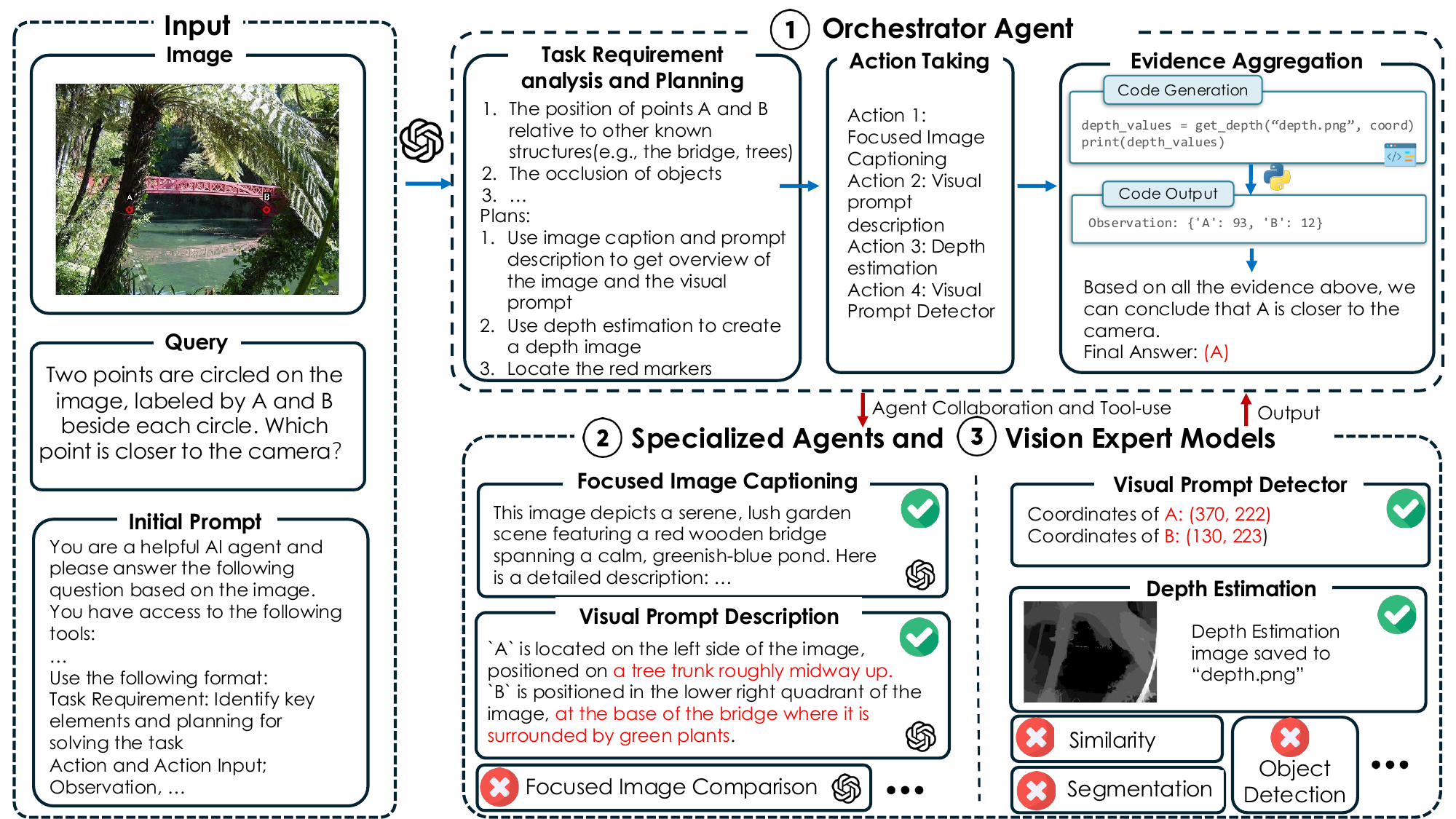}
\caption{
The \textsc{VipAct} framework for visual perception. It consists of (1) an orchestrator agent for task analysis and coordination, (2) specialized agents for specialized visual analysis, and (3) vision expert models for providing pixel-level visual information. Note that not all agents and expert models are invoked in every instance. For complete task-solving processes of \textsc{VipAct}, refer to the case studies in Appendix \ref{sec: case_study}.
} 
\label{fig: framework}
\end{figure*}

Our proposal, \textbf{\textsc{VipAct}}, is illustrated in Figure \ref{fig: framework}. It consists of three main components: (1) \textbf{orchestrator agent} (Section \ref{sec: ochestrator}), which controls the entire workflow by analyzing task requirements and task plans, initiating collaboration with other agents, selecting appropriate vision expert models, summarizing evidence from other agents or tools, and deducing the final answer. (2) \textbf{specialized agents} (Section \ref{sec: specialized_agent}), designed to handle specific tasks such as image captioning, visual prompt description, and image comparison. These agents provide detailed information to the orchestrator agent. (3) \textbf{vision expert models} (Section \ref{sec: expert_model}), which include specialized task-specific vision models that provide accurate, fine-grained perceptual information, addressing limitations of current VLMs. Intuitively, \textsc{VipAct} enhances the VLM's System-2 reasoning by generating detailed intermediate reasoning steps through multi-agent collaboration while leveraging the high-precision perceptual information from vision expert models.

\subsection{Orchestrator Agent} \label{sec: ochestrator}
\textbf{Task Requirement Analysis and Planning:} Inspired by recent works \citep{yao2022react, huang2022language, yang2023mm, sun-etal-2024-pearl} that integrate reasoning, planning, and action in LLM-based agent frameworks, the orchestrator agent begins by analyzing the task requirements derived from the images and queries. This analysis identifies the key elements necessary to solve the problem and the corresponding critical visual features that must be acquired in subsequent steps of the agent's workflow, as well as other criteria derived from its own knowledge. The orchestrator agent then generates a detailed plan for tackling the task, outlining the concrete steps required to obtain the necessary information to meet the objectives. For instance, in a depth estimation task as illustrated in Figure \ref{fig: framework}, the orchestrator agent would determine the essential requirements for comparing depth, such as identifying the specific objects targeted by the red circles and recognizing their relative positions to the camera.

\textbf{Tool Selection and Incorporation of Specialized Agents:} After analyzing the task requirements and formulating a plan, the orchestrator agent selects the appropriate tools and specialized agents to provide the visual information necessary to solve the task. Depending on the nature of the task, this may involve initiating collaboration with specialized agents or external vision expert models to gather fine-grained information. Details on these specialized agents and external vision expert models are provided in Sections \ref{sec: specialized_agent} and \ref{sec: expert_model}.

\textbf{Evidence Summarization:} Once the tools and specialized agents have performed their respective tasks in separate environments, the orchestrator agent compiles and summarizes the collected evidence. This involves integrating the outputs from various tools and agents, ensuring that all relevant information is coherently synthesized to support the decision-making process. The orchestrator agent also resolves conflicting evidence and double-checks the factuality of the information, as errors or hallucinations may arise from the expert models and specialized agents.

\textbf{Final Answer Deduction:} With the summarized evidence, the orchestrator agent deduces the final answer. It applies reasoning based on the accumulated information to arrive at an unambiguous conclusion. Depending on the nature and format of the gathered data, the orchestrator agent may generate \texttt{Python} code, which is then executed by an external \texttt{Python} interpreter to derive the final answer. If the gathered information does not lead to a perfect answer, the orchestrator agent is designed to select the closest possible option based on the evidence, supplemented by its own understanding.

\subsection{Collaboration with Specialized Agents} \label{sec: specialized_agent}

\textsc{VipAct} incorporates three specialized agents to enhance its visual perception capabilities: focused image captioning, visual prompt description, and focused image comparison. These agents provide task-specific, detailed information to the orchestrator agent through function calling in a separate environment, integrating their outputs into the main reasoning process. The three specialized agents are described below.

\textbf{Focused Image Captioning:} This agent generates detailed image descriptions, optionally emphasizing specific elements relevant to the task by specifying a \texttt{focus} argument. The \texttt{focus} argument allows for targeted analysis, ranging from general descriptions to particular aspects like "a red car and the background buildings." This flexibility enables the orchestrator agent to obtain precise, task-relevant information from images. Empirical evidence demonstrates its effectiveness across various tasks, with the focus parameter providing fine-grained control over the generated descriptions.

\textbf{Visual Prompt Description:} Specializing in analyzing visual prompts within images (e.g., colored circles, bounding boxes, arrows, textual labels), this agent is crucial for interpreting visual annotations. It generates detailed descriptions of these elements, including their locations, characteristics, and most importantly, the regions or objects these visual prompts target. This enables the orchestrator agent to accurately interpret highlighted or annotated image sections. The agent has shown particular efficacy in tasks involving images with visual prompts, significantly enhancing the system's ability to understand and reason about annotated visual data. 

\textbf{Focused Image Comparison:} This agent analyzes multiple images, identifying similarities and differences with an optional focus on specific elements. Similarly, the \texttt{focus} parameter allows for targeted comparative analysis, either generally or on specific features as directed by the orchestrator agent. This function can provide a detailed comparison of orientations of objects which can be useful in tasks such as multi-view reasoning. This capability is valuable for tasks requiring multi-image input, such as change detection or pattern identification across images. Empirical results demonstrate this agent's exceptional effectiveness in tasks involving multiple image inputs, with the focus parameter enabling precise comparative analyses.

The prompts for these agents are in Appendix \ref{sec: prompt}. \textsc{VipAct} decomposes complex visual tasks into sub-tasks handled by specialized agents, with an orchestrator agent integrating their outputs. The architecture is extensible, allowing for the addition of new agents to address emerging tasks.

\subsection{Integration of Vision-Expert Models}
\label{sec: expert_model}
\textsc{VipAct} further enhances its visual perception capabilities by integrating a suite of vision-expert models, each specializing in specific aspects of image analysis. They collaborate with the orchestrator agent through function calling, uniquely returning both textual data and processed images—making \textsc{VipAct} among the earliest agent frameworks that incorporate \textbf{visual information directly into the reasoning workflow}. These vision-expert models provide fine-grained visual perception information that is often lacking in current VLM's pre-training data \citep{zhang2024llava}. The vision expert tools used in our experiments are described below: 

\textbf{Visual Prompt Detector:} Identifies and localizes annotated elements in images, such as circles, bounding boxes, or other highlighted regions. This tool is crucial for understanding visual instructions or annotations, enabling the agent to focus on relevant areas for analysis. It returns the coordinates of these visual prompts, which often serve as intermediate information to achieve the final answer. 

\textbf{Depth Estimator:} Analyzes spatial relationships within scenes, providing crucial information about the relative distances of objects from the camera. This tool enhances the agent's understanding of 3D structure in 2D images, vital for spatial reasoning tasks. It returns a grey-scale depth image that can be directly input into the orchestrator agent, allowing it to interpret depth information or combine it with other evidence to reach the final answer. 

\textbf{Object Detection:} Identifies and localizes objects within an image, providing the agent with a comprehensive inventory of visible objects, their locations, and sizes. This facilitates detailed scene understanding and object-centric reasoning. The tool returns both a processed image with detected objects' bounding boxes and textual information about these bounding boxes and objects.

\textbf{Image Segmentation:} Offers precise delineation of image regions, separating objects, backgrounds, and distinct areas. This enables fine-grained analysis of image components, crucial for tasks requiring detailed understanding of object boundaries and spatial relationships. It returns images with segmentation masks along with textual information.

\textbf{Embedding-based Similarity Computation:} Quantifies visual similarities across images by generating compact representations of visual content. This allows for nuanced comparisons and similarity assessments, particularly useful for tasks involving image retrieval or comparative analysis. It returns similarity scores based on the selected embedding model and specified similarity metrics, such as cosine similarity. 

The complete function heads, including inputs, outputs, and descriptions for these vision expert models, are provided in the initial prompt for the orchestrator agents in Appendix \ref{sec: prompt}. This diverse toolkit empowers the orchestrator agent to select the most appropriate tools for each task dynamically, significantly enhancing the framework's ability to comprehend and reason about complex visual scenarios. The integration of processed images alongside textual outputs in the agent's workflow enables more nuanced and contextually rich visual reasoning. We provide an overview of the VipAct framework in Algorithm~\ref{alg:vipact-framework} with detailed explanations in Appendix \ref{sec:func_in_algorithm}.

\begin{algorithm*}
\caption{\textsc{VipAct}: \!\textbf{\underline{VI}}sual-\textbf{\underline{P}}erception 
via VLM \textbf{\underline{A}}gent \textbf{\underline{C}}ollaboration \& \textbf{\underline{T}}ool-use}
\label{alg:vipact-framework}
\begin{algorithmic}[1]
\Require 
Set of visual inputs $\mathcal{V}$, 
a query $q$,
a vision-language model $\mathcal{M}$,
a set of tools $\mathcal{T} = \{T_1, \ldots, T_n\}$ including specialized agents and vision expert models, 
and the maximum iterations $K$
\Ensure An answer $a$ to the visual perception task
\State Initialize orchestrator agent $\mathcal{O}$ with $\mathcal{M}$ and $\mathcal{T}$
\State $\mathcal{P}_0 \gets \textsc{FormatPrompt}(\mathcal{V}, q)$ \Comment{Format initial prompt with visual inputs and query}
\State $t \gets 0$, \; $\mathcal{S} \gets \emptyset$ \Comment{Initialize iteration counter and state}
\While{$t < K$ and not \textsc{IsTerminated}($\mathcal{S}$)}
    \If{$\exists T_i \in \mathcal{T} : \textsc{IsRequired}(T_i, \mathcal{S})$} \Comment{Check if any tool is required}
        \State $T^* \gets \argmax_{T_i \in \mathcal{T}} \textsc{Utility}(T_i, \mathcal{S})$ \Comment{Select most useful tool}
        \State $\mathcal{O}_t \gets \textsc{Execute}(T^*, \mathcal{S})$ \Comment{Execute selected tool with the current state as input}
        \If{$\textsc{ContainsVisualData}(\mathcal{O}_t)$}
            \State $\mathcal{V} \gets \mathcal{V} \cup \textsc{ProcessVisualData}(\mathcal{O}_t)$ \Comment{Add new visual data if needed}
        \EndIf
    \Else
        \State $\mathcal{R}_t \gets \mathcal{M}(\mathcal{P}_{t-1})$ \Comment{Generate VLM output}
        \State $\mathcal{O}_t \gets \textsc{InterpretOutput}(\mathcal{R}_t)$ \Comment{Interpret VLM output}
    \EndIf
    
    \State $\mathcal{P}_t \gets \textsc{UpdatePrompt}(\mathcal{P}_{t-1}, \mathcal{O}_t)$ \Comment{Update prompt with new information}
    \State $\mathcal{S} \gets \textsc{UpdateState}(\mathcal{S}, \mathcal{O}_t)$; $t \gets t + 1$ \Comment{Update state with new observations}
\EndWhile
\State $a \gets \textsc{ExtractAnswer}(\mathcal{S})$ \Comment{Extract final answer from state}
\State \Return $a$

\end{algorithmic}
\end{algorithm*}

\section{Experiment}

\textbf{Setup.} We use various SOTA closed-source VLMs, including \textbf{GPT-4o} \citep{gpt4o}, \textbf{Gemini-1.5-Pro} \citep{team2024gemini}, and \textbf{Claude-3.5-Sonnet} \citep{anthropic_claude3_5_sonnet}, as well as open-source ones, such as \textbf{LLaVA-OneVision-7B} \citep{li2024llava}, \textbf{InternVL-2-Pro} \citep{chen2023internvl, chen2024far}, and \textbf{Llama-3.2-90b-Vision} \citep{dubey2024llama}. Following prior works \citep{zheng2024gpt, he-etal-2024-webvoyager, liu2024visualwebbench, gu2024your}, we focus on GPT-4o \citep{gpt4o} as the primary VLM for analysis in the main paper due to page constraints. Discussions on other VLMs are included in Appendix \ref{sec: other_vlms}, with additional implementation details in Appendix \ref{sec:implement}.

\textbf{Datasets.} To evaluate VLMs on visual perception tasks, we use the following two challenging datasets designed to test fine-grained visual perception: (1) \textbf{Blink} \citep{fu2024Blink} includes diverse visual tasks solvable by humans ``within a blink,'' yet difficult for SOTA VLMs. It features visual prompts such as bounding boxes and interleaved image-text formats, often with multiple images in a single query. We use Blink as the main benchmark. (2) \textbf{MMVP} \citep{Tong_2024_CVPR} is a benchmark for evaluating visual grounding in VLMs, using image pairs from ``CLIP-blind pairs''—visually distinct images that are similar in CLIP embedding space. It focuses on nine basic visual patterns that are easy for humans but challenging for SOTA VLMs. 
Details are in Appendix \ref{sec:dataset}. 

\textbf{Baselines.} We evaluate \textsc{VipAct} against four types of baselines:
(1) \textbf{Text-based prompting}, including zero-shot prompting; chain-of-thought (CoT) prompting \citep{wei2022chain, kojima2022large}; Least-to-most prompting (LtM) \citep{zhou2022least}; and Tree-of-thought (ToT) prompting \citep{yao2024tree}.
(2) \textbf{Few-shot in-context learning} \cite{brown2020language}, where in-context exemplars are selected using different strategies, including random selection, or selection based on embedding \citep{radford2021learning, dosovitskiy2020image} similarity (analyzed separately in Appendix \ref{sec: few-shot}).
(3) \textbf{Visual Prompting}, exemplified by Set-of-Mark (SoM) \citep{yang2023set}, which overlays marks on semantically meaningful image regions.
(4) \textbf{Vision language agentic frameworks}, including MM-ReAct \citep{yang2023mm}, which integrates LLMs with vision experts via ReAct-style prompts \citep{yao2022react}; ViperGPT \citep{suris2023vipergpt}, using LLMs to generate code composing vision and language models; and VisProg \citep{gupta2023visual}, which generates visual programs from textual instructions.

\begin{table*}[!t]
  \centering
  % Use a smaller font size for the table
  \small
  % Reduce column spacing to fit the table within the text width
  \setlength{\tabcolsep}{7pt} 
  \begin{tabular}{lccccccccccc}
    \toprule
    % Abbreviated column headers to save space
    \textbf{Method} & \textbf{Sim} & \textbf{Count} & \textbf{Depth} & \textbf{Jig} & \textbf{Fun.C} & \textbf{Sem.C} & \textbf{Spat} & \textbf{Local} & \textbf{Vis.C} & \textbf{Multi-v} & \textbf{Average} \\
    \midrule
    % Converted row color to CMYK format as in the example provided
    \rowcolor[cmyk]{0,0,0,0.07} 
    % Corrected typo in the sub-header
    \multicolumn{12}{l}{\textit{Text-based Prompting w/ GPT-4o}} \\
    Zero-shot \cellcolor{green!10} & 65.44 & 50.83 & 64.52 & 60.00 & 57.69 & 56.83 & 79.92 & 56.00 & 86.05 & 60.15 & 63.74 \\
    CoT \cellcolor{green!10} & 63.70 & 65.00 & 73.39 & 62.00 & 57.69 & 57.55 & 82.52 & 60.66 & 82.56 & 53.38 & 65.85 \\
    LtM \cellcolor{green!10} & 62.22 & 64.17 & 70.97 & 62.67 & 55.38 & 55.40 & 76.22 & 59.02 & 83.14 & 45.86 & 63.51 \\
    ToT \cellcolor{green!10} & 64.44 & 58.33 & 71.70 & 64.00 & 57.69 & 59.71 & 83.22 & 61.48 & 78.49 & 50.38 & 64.94 \\
    \rowcolor[cmyk]{0,0,0,0.07} 
    \multicolumn{12}{l}{\textit{Visual Prompting w/ GPT-4o}} \\
    SoM \cellcolor{blue!10} & 63.70 & 43.33 & 68.55 & 49.33 & 47.69 & 52.52 & 76.22 & 59.84 & 83.72 & 56.40 & 60.13 \\
    \rowcolor[cmyk]{0,0,0,0.07} 
    \multicolumn{12}{l}{\textit{Multi-modal Agent Framework w/ GPT-4o}} \\
    MM-ReAcT \cellcolor{cyan!10} & - & 30.00 & 0.81 & - & - & - & 63.64 & 0.00 & - & - & - \\
    ViperGPT \cellcolor{cyan!10} & - & 29.17 & 0.00 & - & - & - & 48.95 & 18.85 & - & - & - \\
    VisProg \cellcolor{cyan!10} & - & 3.33 & 0.00 & - & - & - & 31.47 & 14.75 & - & - & - \\
    \textbf{\textsc{VipAct} (Ours)}\cellcolor{cyan!10} & \textbf{81.48} & \textbf{70.00} & \textbf{90.80} & \textbf{68.00} & \textbf{61.50} & \textbf{60.40} & \textbf{86.70} & \textbf{63.11} & \textbf{91.28} & \textbf{62.63} & \textbf{73.79} \\
    \bottomrule
  \end{tabular}
  \caption{Results for visual reasoning tasks in Blink using GPT-4o. Note that ``$-$'' indicates methods that do not support multiple images. Our \textsc{VipAct} consistently outperforms baselines on all tasks.}
  \label{tab: main_result}
\end{table*}

\textbf{Result Analysis.} Tables \ref{tab: main_result} and \ref{tab:result_mmvp_gpt4o} present the performance of our proposed \textsc{VipAct} framework and baseline methods on each sub-task of the Blink and MMVP datasets respectively. We make the following key observations:
\textbf{(1) Text-based prompting methods do not consistently improve performance over zero-shot prompting.} Specifically, as shown in Tables \ref{tab: main_result} and \ref{tab:result_mmvp_gpt4o}, prior text-based prompting methods effective for LLMs — such as CoT — can improve performance on some sub-tasks like visual similarity, object localization, counting, and spatial relations. However, for other tasks, the improvement is minimal or even negative. More advanced techniques like LtM and ToT exhibit similar phenomena. Empirically, while these methods elicit detailed reasoning, such steps are often ungrounded in visual elements and can cause severe hallucinations. Therefore, it is non-trivial to elicit VLMs' reasoning for better general visual perception using text-based methods from text-only LLMs.
\textbf{(2) SoM can impair VLMs' fine-grained perception in most scenarios.} From results on both datasets, SoM adversely affects VLM performance on almost all tasks. Empirically, overlaying labeled masks can become cluttered with numerous semantic objects or fine-grained parts, negatively influencing VLM perception of original objects and potentially confusing models with original visual prompts and labels. Consequently, SoM's effectiveness in some compositional reasoning tasks with limited semantic objects does not generalize well to broader visual perception tasks, especially those requiring visual prompt understanding.
\textbf{(3) Previous visual programming methods exhibit poor generalization ability.} As shown, these methods perform adequately only on limited tasks (e.g., spatial relations, counting) similar to those in common VQA datasets \citep{hudson2019gqa, suhr-etal-2019-corpus, marino2019ok}. Their generated code calls a limited set of predefined tools, lacking logic for unsupported scenarios or errors. They cannot support images with visual prompts, failing to locate them for subsequent operations (e.g., near-zero performance in depth estimation due to inability to locate red circles, leading to non-executable code). Moreover, code generated solely from text queries lacks flexibility for different image characteristics. These observations highlight the need for a generalizable agent framework leveraging both vision expert models and VLM flexibility.
\textbf{(4) \textsc{VipAct} consistently achieves the best performance across all sub-tasks in Blink and MMVP, demonstrating its effectiveness and generalization ability.} By examining \textsc{VipAct}'s reasoning traces, we observe that, compared to text-based and visual prompting methods, \textsc{VipAct} effectively invokes specialized agents or vision expert models to enhance image understanding. It does not solely rely on their outputs, as evidence might be incorrect or errors may occur. Instead, it aggregates useful evidence with additional reasoning to infer the final answer, showcasing its ability to handle uncertainties and integrate multiple information sources. Figure \ref{fig: case_study_1} and \ref{fig: case_study_2} in Appendix \ref{sec: case_study} show complete reasoning traces of \textsc{VipAct}.

\begin{table}[h]
\centering
\begin{tabular}{@{}lc@{}}
\toprule
\textbf{Method} & \textbf{Accuracy (\%)} \\
\midrule
Zero-shot & 68.0 \\
CoT       & 61.0 \\
LtM       & 66.0 \\
ToT       & 66.0 \\
SoM       & 62.0 \\
MM-ReAct  & 6.67 \\
ViperGPT  & 53.0 \\
VisPro    & 39.0 \\
\textbf{\textsc{VipAct} (Ours)} & \textbf{70.7} \\
\bottomrule
\end{tabular}
\caption{Different methods using GPT-4o on MMVP.}
\label{tab:result_mmvp_gpt4o}
\vspace{-0.24in}
\end{table}

\section{Ablation Study}

To evaluate the effectiveness of various components in \textsc{VipAct}, we conduct a series of ablation studies. These involve removing or modifying key components of \textsc{VipAct} to assess their impact on performance across different tasks. The ablation studies are as follows: \textbf{(1) Removal of multi-agent collaboration}: We removed the specialized agents and incorporated their prompts as instructions directly into the orchestrator agent to evaluate the importance of multi-agent collaboration. \textbf{(2) Removal of image input for orchestrator agent: }We modified the input to the orchestrator agent to only include image paths as text, rather than the actual images which means the image is not visible to the orchestrator agent but still can be served as input for other specialized agents or vision expert models. This setup follows the paradigm used in previous works \citep{suris2023vipergpt, gupta2023visual} and tests the effectiveness of direct visual input to the orchestrator agent. \textbf{(3) Removal of specialized agents:} We removed all specialized agents to assess their impact on the \textsc{VipAct}'s performance. \textbf{(4) Removal of vision expert models: } We eliminated all vision-expert models to evaluate their contribution.

\begin{table*}[!t]
  \centering
  % Reduce column spacing to fit the table within the text width.
  \setlength{\tabcolsep}{7pt} 
  \begin{tabular}{lcccccccccc}
    \toprule
    \textbf{Method} & \textbf{Sim} & \textbf{Count} & \textbf{Depth} & \textbf{Jig} & \textbf{Fun.C} & \textbf{Sem.C} & \textbf{Spat} & \textbf{Local} & \textbf{Vis.C} & \textbf{Multi-v} \\
    \midrule
    % The row color is kept, but converted to the required CMYK format.
    \rowcolor[cmyk]{0,0,0,0.07}
    \multicolumn{11}{l}{\textit{Variants of \textsc{VipAct}}} \\
    \textsc{VipAct} (Full) & \textbf{81.48} & \textbf{70.00} & \textbf{90.80} & \textbf{68.00} & \textbf{61.50} & \textbf{60.40} & \textbf{86.70} & 63.11 & \textbf{91.28} & \textbf{62.63} \\
    \quad w/o Multi-agent & 80.00 & 67.50 & 75.00 & 66.00 & 58.46 & 59.71 & 82.52 & \textbf{63.93} & 85.47 & 48.87 \\
    \quad w/o Visual Input & 77.78 & 59.71 & 69.35 & 61.33 & 53.85 & 51.08 & 83.22 & 60.66 & 78.49 & 48.12 \\
    \quad w/o Spec. Agents & 65.72 & 62.45 & 85.62 & 62.32 & 55.25 & 56.32 & 81.96 & 58.49 & 75.48 & 46.75 \\
    \quad w/o Vision Expert & 64.34 & 57.44 & 72.58 & 65.67 & 59.42 & 58.59 & 81.37 & 57.44 & 83.63 & 56.40\\
    \bottomrule
  \end{tabular}
  \caption{Ablation study results of \textsc{VipAct} on the Blink benchmark using GPT-4o. \textsc{VipAct} (Full) represents the complete framework with all components, while the other variants exclude specific components.}
  \label{tab: ablationd_Blink}
\end{table*}

\begin{table}[h]
\centering
\vspace{-0.2in}
\begin{tabular}{@{}lc@{}}
\toprule
\textbf{Method} & \textbf{Accuracy (\%)} \\
\midrule
\textsc{VipAct} & \textbf{70.7} \\
\quad w/o Multi-agent & 68.0 \\
\quad w/o Visual Input & 54.0 \\
\quad w/o Spec. Agents & 67.0 \\
\quad w/o Vision Expert & 66.0 \\
\bottomrule
\end{tabular}
\caption{Ablation of \textsc{VipAct} on MMVP using GPT-4o.}
\vspace{-0.2in}
\label{tab:ablationd_mmvp}
\end{table}

The results of ablation studies are presented in Table~\ref{tab: ablationd_Blink} and \ref{tab:ablationd_mmvp}. From these results, we derive the following key insights:
\begin{itemize}
    \item \textbf{Multi-agent collaboration enhances detailed reasoning}: The removal of multi-agent collaboration led to a consistent performance decline. By comparing reasoning steps, we observed that multi-agent collaboration enabled significantly more detailed image analysis (over 80\% more generated tokens), such as thorough image captioning. This aligns with observations in LLMs \citep{wu2023autogen, hong2023metagpt, qian2023communicative,10.1145/3586183.3606763,liu2023dynamic}, where agent collaboration enhances task-solving via comprehensive reasoning from diverse perspectives.
    \item \textbf{Direct image input to the orchestrator agent is essential for flexible task planning and error handling}: As shown in Tables~\ref{tab: ablationd_Blink} and \ref{tab:ablationd_mmvp}, removing direct image input to the orchestrator significantly degrades performance. Without direct visual access, the orchestrator agent relies solely on textual queries, lacking critical visual information for accurate task planning. This leads to suboptimal decision-making, less precise parameter selection (e.g., the \texttt{focus} parameter), and overly generalized task analysis, reducing specificity. For instance, in a multi-view reasoning task, direct image input allows the agent to identify reference objects, enabling it to accurately adjust the \texttt{focus} parameter and effectively determine the direction of camera movement. Further analysis is in Appendix \ref{sec:visual_input_additional}.
    \item \textbf{Specialized agents and vision expert models significantly contribute to performance}: Specialized agents, though VLMs, intently analyze specific visual information without distractions from other instructions (e.g., format requirements), which can hinder LLM reasoning \citep{tam2024let}. Vision expert models perform pixel-level analyses beyond SOTA VLM capabilities, aiding the orchestrator. As shown in Table~\ref{tab: ablationd_Blink} and \ref{tab:ablationd_mmvp}, removing them leads to a noticeable performance decline.
Overall, our \textsc{VipAct} framework combines VLM flexibility and planning with vision expert model precision, creating a cohesive system where each component is essential.
\end{itemize}

\section{Error Analysis} \label{sec: error_analysis}

To examine the limitations of GPT-4o's visual perception capabilities and the bottlenecks of our \textsc{VipAct}, we conduct a detailed error analysis. Following prior works \citep{zhou2022least, chen2023skills, zhang2024darg}, we randomly sampled 20 error cases from each sub-task within the two datasets. The errors were categorized as follows: 
\begin{itemize}
    \item \textbf{Failure to perceive small object parts (17\%):} The model often overlooks small, semantically important components crucial for precise visual understanding.
    \item \textbf{Difficulty distinguishing closely positioned visual prompts (15\%):} The model struggles to differentiate spatially proximate visual prompts.
    \item \textbf{Challenges in fine-grained spatial reasoning (24\%):} Tasks requiring high spatial resolution highlight the model's \textbf{bias towards foreground objects over backgrounds}; e.g., misinterpreting a highlight meant for the sky near a car as associated with the car.
    \item \textbf{Misinterpretation of relative object positions (14\%):} Errors arise when object arrangements differ from real-world expectations, as the model often lacks the ability to infer spatial relations from objects' perspectives, focusing on camera viewpoint.
    \item \textbf{Failure to recognize object orientation (13\%):} Difficulty discerning object orientation leads to errors in recognizing object parts, such as distinguishing left/right bicycle pedals.
    \item \textbf{Other errors (17\%):} This includes other issues like failure to detect subtle color differences, inaccuracies in multi-image fine-grained structure correspondence, and instances of refusal or instruction misinterpretation.
\end{itemize}

Case studies illustrating these errors are in Appendix \ref{sec: case_study}. Our analysis denotes that while \textsc{VipAct} shows significant improvements in VLM visual perception, fine-grained perception remains a bottleneck. Specifically, the model lacks the \textbf{spatial intelligence or imaginative abilities} \citep{Chen_2018_CVPR, huang2024rekep} necessary to infer relative object positions beyond their pixel positions (from the camera's perspective), especially in the context of real-life scenes. Noticeably, these limitations hinder the model's ability to accurately interpret visual prompts and process tasks involving multiple image inputs. We also examine the significance of multiple image inputs for VLMs in Appendix \ref{sec: multiple_image}.

\section{Conclusion}
We introduce \textbf{\textsc{VipAct}}, a VLM-based agent framework that synergizes multi-agent collaboration and vision expert models for fine-grained visual perception tasks. By combining the planning and function-calling capabilities of SOTA VLMs, \textsc{VipAct} enhances VLMs' System-2 reasoning through multi-agent interactions and integrates high-precision, pixel-level information from specialized vision models. Our experiments across a diverse range of visual perception tasks demonstrate that \textsc{VipAct} achieves SOTA performance, outperforming previous baselines. The comprehensive ablation study highlights the critical role of multi-agent collaboration in eliciting detailed information for reasoning, as well as the importance of image input in task planning. Furthermore, our error analysis highlights several inherent limitations in current SOTA VLMs that form bottlenecks in our framework, offering valuable insights for future improvements.
    \bibliography{aaai2026}

\clearpage
\appendix

\section{Implementation Details} \label{sec:implement}

For main experiments, we use the \texttt{gpt-4o-2024-05-13} model from Azure OpenAI API. Following previous works \citep{fu2024Blink} to ensure reproducibility, we set the temperature to 0 for all VLM inference and set the maximum number of tokens to 2048. For components of \textsc{VipAct}, we use the same \texttt{gpt-4o-2024-05-13} model for the implementation of orchestrator agents and specialized agents. For the implementation of vision expert models, we use the \texttt{Depth-Anything-V2-Small-hf} checkpoint \citep{yang2024depth} for depth estimation, the Segment Anything Model (SAM) \citep{kirillov2023segment} for segmentation, the YOLOv8 model \citep{hussain2023yolo} from Ultralytics for object detection, and the \texttt{clip-vit-base-patch32} \citep{radford2021learning} for similarity comparison using cosine similarity.
For experiments with LLaVA, we use the latest SOTA \texttt{llava-onevision-qwen2-7b-ov} \citep{li2024llava}, which is one of the few VLMs that support multiple images as inputs and achieves SOTA results on various vision-language benchmarks \citep{li2024llava2, bansal2020visual} compared to other open-source models of similar size. For the implementation of all prompting baselines, we adopt the codebase from the original Blink and MMVP papers and use the exact same settings, including the method for computing performance. For the implementation of baselines MM-ReAct, ViperGPT, and VisProg, we adopt the original codebase they provide, except that the backbone model is replaced with GPT-4o, as their original models such as Codex \citep{chen2021evaluating} are not available and to ensure fair comparison. For the implementation of few-shot in-context learning, the embedding models' checkpoints we use are \texttt{clip-vit-base-patch32} and \texttt{vit-base-patch16-224} \citep{alexey2020image}.
For all experiments, we run three times and report the average number. For the results in Table \ref{tab: main_result} and \ref{tab:result_mmvp_gpt4o}, we conduct significance tests following \cite{berg-kirkpatrick-etal-2012-empirical}. The average estimate of p-value is 0.006 ($<$ 0.01) between \textsc{VipAct} and SOTA baselines, demonstrating significant differences.
The total inference time for our \textsc{VipAct} on Blink and MMVP is less than 2 hours, which is acceptable. Our computational resources consist of a Linux server with 4 NVIDIA A100-40G GPUs.

\section{Dataset Details} \label{sec:dataset}

In this section, we provide the details of the dataset used in our experiments. The Blink \citep{fu2024Blink} dataset contains a variety of tasks that evaluate different aspects of VLMs' perception capabilities. In our paper, we specifically focus on the following sub-tasks: Similarity (\textbf{Sim}), Counting (\textbf{Count}), Depth Estimation (\textbf{Depth}), Jigsaw Puzzle (\textbf{Jig}), Functional Correspondence (\textbf{Fun.C}), Semantic Correspondence (\textbf{Sem.C}), Spatial relation (\textbf{Spat}), Local Correspondence (\textbf{Local}), Visual Correspondence (\textbf{Vis.C}), and Multi-view Reasoning (\textbf{Multi-v}). The dataset is divided into validation and test sets, with the number of data points for each sub-task as shown in Table~\ref{tab: dataset_statistic}.

\begin{table*}[htb]
\centering
% Use tabular* to make the table span the full text width.
% The @{\extracolsep{\fill}} command distributes space between columns.
\begin{tabular*}{\textwidth}{l @{\extracolsep{\fill}} cccccccccc}
\toprule
\textbf{Sub-task} & \textbf{Sim} & \textbf{Count} & \textbf{Depth} & \textbf{Jig} & \textbf{Fun.C} & \textbf{Sem.C} & \textbf{Spat} & \textbf{Local} & \textbf{Vis.C} & \textbf{Multi-v} \\
\midrule
\textbf{Validation} & 135 & 120 & 124 & 150 & 130 & 139 & 143 & 122 & 172 & 133 \\
\textbf{Test} & 136 & 120 & 124 & 150 & 130 & 140 & 143 & 125 & 172 & 133 \\
\bottomrule
\end{tabular*}
\caption{Number of data points for each sub-task in the validation and test sets of Blink.}
\label{tab: dataset_statistic}
\end{table*}

The tasks and the corresponding datasets are described in the original Blink paper. Each sub-task is designed to challenge different aspects of the model's perceptual reasoning capabilities, as detailed in the main text of our paper. Following previous works \citep{hu2024visual}, we exclude datasets focused on compositional reasoning like IQ testing or commonsense reasoning, as they do not directly assess visual perception and more focus on compositional reasoning.

Another dataset we use in this work is the Multimodal Visual Patterns (MMVP) dataset \citep{Tong_2024_CVPR} which consists of 150 CLIP-blind image pairs and 300 associated visual questions, designed to probe nine core visual patterns: orientation, presence of specific features, state, quantity, positional context, color, structure, text, and viewpoint. Human participants achieved 95.7\% accuracy, while state-of-the-art MLLMs, including GPT-4V and Gemini, performed significantly worse. The dataset highlights fundamental failures in visual grounding tasks and serves as a benchmark for advancing VLMs' visual perception ability.

\section{Exploration of different VLMs} \label{sec: other_vlms}

In addition to the GPT-4o used in our main experiments, we also evaluate other VLMs on our tasks. Specifically, we explore five additional SOTA VLMs, including (1) open-source models: LLaVA-OneVision-7B \citep{li2024llava}, the latest open-source model in the LLaVA series, InternVL-2-Pro \citep{chen2023internvl, chen2024far}, and Llama-3.2-90b-vision \citep{dubey2024llama}; and (2) close-source models: Gemini-1.5-Pro \citep{team2024gemini} and Claude-3.5-Sonnet \citep{anthropic_claude3_5_sonnet}. 

For open-source models, we find that applying \textsc{VipAct}'s prompt (described in Section \ref{sec: prompt}) reveals significant limitations. These VLMs often fail to follow key instructions, such as adhering to the required format, which is critical for extracting the tool-use indicators necessary for integrating external tools. Furthermore, they frequently generate image captions even when no such instruction is provided, suggesting a bias towards image captioning or description tasks. 

To evaluate these open-source VLMs comprehensively, we apply prompting baselines and report the results on the Blink benchmark and MMVP in Table \ref{tab: Blink_result_llava} and Table \ref{tab:result_open_mmvp}. These results demonstrate that while LLaVA-OneVision-7B achieves above-random accuracy on tasks like object counting and spatial relations—typical of standard VQA problems found in prior datasets \citep{li2024llava2, bansal2020visual}—it performs near or below random on other tasks. We also observe significant positional biases \citep{zhang2024position, shi2024judging}, with this model frequently predicting the first option for most data points within a task. In contrast, InternVL-2-Pro and Llama-3.2-90b-vision exhibit better performance, though still significantly behind GPT-4o. These findings indicate that current open-source SOTA VLMs struggle with generalizing to more complex or non-standard VQA tasks, lacking the fine-grained perception capabilities necessary for broader applicability. Moreover, alternative prompting strategies do not yield noticeable improvements over the zero-shot baseline for these models.

In contrast, the two additional close-source VLMs—Gemini-1.5-Pro and Claude-3.5-Sonnet—demonstrate instruction-following abilities comparable to GPT-4o, allowing effective application of our \textsc{VipAct} framework. As shown in Tables \ref{tab:gemini_claude_result}, applying \textsc{VipAct} on these models consistently outperforms previous prompting baselines, achieving significant improvements. These results highlight the effectiveness and generalization capability of our approach when used with models possessing strong instruction-following capabilities.

\begin{table*}[!t]
\centering
\setlength{\tabcolsep}{7pt} % Manually set the space between columns. Adjust as needed.
\begin{tabular}{lccccccccccc}
    \toprule
    \textbf{Method} & \textbf{Sim} & \textbf{Count} & \textbf{Depth} & \textbf{Jig} & \textbf{Fun.C} & \textbf{Sem.C} & \textbf{Spat} & \textbf{Local} & \textbf{Vis.C} & \textbf{Multi-v} & \textbf{Overall} \\
    \midrule
    \rowcolor[rgb]{0.93,0.93,0.93} 
    \multicolumn{12}{l}{\textit{Text-based Prompting w/ Gemini-1.5-Pro}} \\
    Zero-shot \cellcolor{green!10} & 78.52 & 60.83 & 70.97 & 72.67 & 44.62 & 51.08 & 74.13 & 57.38 & 81.98 & 55.64 & 64.78 \\
    CoT \cellcolor{green!10} & 81.48 & \textbf{64.17} & 78.23 & 68.67 & 42.31 & 53.96 & 78.32 & 60.66 & 81.98 & 51.88 & 66.17 \\
    LtM \cellcolor{green!10} & 84.44 & 62.50 & 75.81 & 73.33 & 46.92 & 50.36 & 73.43 & 63.11 & 84.30 & 54.14 & 66.83 \\
    ToT \cellcolor{green!10} & 82.23 & 61.52 & 74.84 & 68.96 & 45.42 & 52.61 & 75.34 & 61.25 & 82.21 & 52.86 & 65.72 \\
    \rowcolor[rgb]{0.93,0.93,0.93} 
    \multicolumn{12}{l}{\textit{Visual Prompting w/ Gemini-1.5-Pro}} \\
    SoM \cellcolor{blue!10} & 60.74 & 55.00 & 65.32 & 62.00 & 47.69 & 43.88 & 74.13 & 59.02 & 74.42 & 53.38 & 59.56 \\
    \rowcolor[rgb]{0.93,0.93,0.93} 
    \multicolumn{12}{l}{\textit{Multi-modal Agent Framework w/ Gemini-1.5-Pro}} \\
    VipAct \cellcolor{cyan!10} & \textbf{84.44} & \textbf{64.17} & \textbf{89.42} & \textbf{74.00} & \textbf{48.74} & \textbf{57.55} & \textbf{79.57} & \textbf{70.48} & \textbf{86.05} & \textbf{59.42} & \textbf{71.38} \\
    \midrule
    \rowcolor[rgb]{0.93,0.93,0.93} 
    \multicolumn{12}{l}{\textit{Text-based Prompting w/ Claude-3.5-Sonnet}} \\
    Zero-shot \cellcolor{green!10} & 85.19 & 67.50 & 66.13 & 58.00 & 58.00 & 44.60 & 72.03 & 57.38 & 73.84 & 48.12 & 63.08 \\
    CoT \cellcolor{green!10} & 86.72 & \textbf{68.33} & 71.77 & 61.33 & 52.31 & 41.73 & 77.62 & 50.00 & 81.98 & 44.36 & 63.62 \\
    LtM \cellcolor{green!10} & 87.42 & 67.42 & 68.42 & 59.97 & 58.00 & 45.13 & 73.82 & 57.47 & 74.29 & 47.86 & 63.98 \\
    ToT \cellcolor{green!10} & 86.90 & 67.53 & 69.48 & 57.35 & 59.46 & 43.72 & 74.92 & 58.49 & 76.14 & 46.38 & 64.04 \\
    \rowcolor[rgb]{0.93,0.93,0.93} 
    \multicolumn{12}{l}{\textit{Visual Prompting w/ Claude-3.5-Sonnet}} \\
    SoM \cellcolor{blue!10} & 82.65 & 62.78 & 63.81 & 56.79 & 56.73 & 39.58 & 72.00 & 52.47 & 73.74 & 44.63 & 60.52 \\
    \rowcolor[rgb]{0.93,0.93,0.93} 
    \multicolumn{12}{l}{\textit{Multi-modal Agent Framework w/ Claude-3.5-Sonnet}} \\
    VipAct \cellcolor{cyan!10} & \textbf{88.89} & 67.96 & \textbf{88.59} & \textbf{65.33} & \textbf{60.42} & \textbf{50.13} & \textbf{78.82} & \textbf{61.54} & \textbf{83.72} & \textbf{49.57} & \textbf{69.50} \\
    \bottomrule
\end{tabular}
\caption{Results for visual reasoning tasks in Blink using Gemini-1.5-Pro and Claude-3.5-Sonnet. Our \texttt{VipAct} consistently outperforms baselines on almost all tasks.}
\label{tab:gemini_claude_result}
\end{table*}

\begin{table*}[!t]
  \centering
  \begin{tabular}{lccc}
    \toprule
    \textbf{Method} & \textbf{LLaVA-OneVision-7B} & \textbf{InternVL-2-Pro} & \textbf{Llama-3.2-90b-vision} \\
    \midrule
    Random & 25.00 & 25.00 & 25.00 \\
    Zero-shot & 29.67 & 60.00 & 57.33 \\
    CoT & 30.33 & 57.33 & 59.33 \\
    LtM & 30.00 & 58.67 & 57.33 \\
    ToT & 31.33 & 60.00 & 59.38 \\
    SoM & 27.00 & 45.33 & 51.33 \\
    \bottomrule
  \end{tabular}
  \caption{Results of different open-source VLMs with different prompting methods on the MMVP benchmark, including a random baseline for comparison.}
  \label{tab:result_open_mmvp}
\end{table*}

\begin{table*}[!t]
  \centering
  \begin{tabular}{lcccccccccc}
    \toprule
    \textbf{Method} & \textbf{Sim} & \textbf{Count} & \textbf{Depth} & \textbf{Jig} & \textbf{Fun.C} & \textbf{Sem.C} & \textbf{Spat} & \textbf{Local} & \textbf{Vis.C} & \textbf{Multi-v} \\
    \midrule
    Random & 50.00 & 25.00 & 50.00 & 50.00 & 25.00 & 25.00 & 50.00 & 50.00 & 25.00 & 50.00 \\
    \midrule
    Zero-shot & 47.41 & 63.33 & 51.61 & 52.67 & 20.00 & 23.02 & 72.73 & 50.82 & 23.26 & 44.36 \\
    CoT  & 44.44 & 57.20 & 54.03 & 52.67 & 20.77 & 25.90 & 76.22 & 43.44 & 22.67 & 35.34 \\
    LtM  & 45.93 & 56.67 & 51.61 & 52.67 & 15.38 & 28.87 & 72.03 & 50.82 & 30.81 & 42.11 \\
    ToT  & 47.41 & 63.33 & 50.00 & 52.67 & 15.38 & 24.46 & 72.03 & 50.82 & 23.26 & 44.36 \\
    SoM & 47.41 & 46.67 & 54.03 & 52.67 & 23.85 & 21.58 & 72.73 & 41.80 & 19.19 & 31.58 \\
    \bottomrule
  \end{tabular}
  \caption{Result of baseline methods evaluated using LLaVa-OneVision-7B on the Blink dataset.}
  \label{tab: Blink_result_llava}
\end{table*}

\begin{table}[htbp]
  \centering
  \begin{tabular}{lcc}
    \toprule
    \multirow{2}{*}{Dataset} & \multicolumn{2}{c}{Model} \\
    \cmidrule(lr){2-3}
    & GPT-4o & LLaVA-OneVision-7B \\
    \midrule
    Sim & 59.51 \textcolor{darkred}{($\downarrow$ -5.93)} & 45.93 \textcolor{darkred}{($\downarrow$ -1.48)}\\
    Jig & 57.78 \textcolor{darkred}{($\downarrow$ -2.22)} & 52.67 \textcolor{gray}{($\rightarrow$ 0.00)}\\
    Fun.C & 53.34 \textcolor{darkred}{($\downarrow$ -4.35)} & 20.00 \textcolor{gray}{($\rightarrow$ 0.00)}\\
    Sem.C & 56.60 \textcolor{darkred}{($\downarrow$ -0.23)} & 24.46 \textcolor{darkgreen}{($\uparrow$ +1.44)}\\
    Vis.C & 83.91 \textcolor{darkred}{($\downarrow$ -2.14)} & 18.60 \textcolor{darkred}{($\downarrow$ -4.66)}\\
    Multi-v & 51.38 \textcolor{darkred}{($\downarrow$ -8.77)} & 29.32 \textcolor{darkred}{($\downarrow$ -15.04)}\\
    \textbf{Overall} & 60.42 \textcolor{darkred}{($\downarrow$ -3.94)} & 31.83 \textcolor{darkred}{($\downarrow$ -3.29)}\\
    \bottomrule
  \end{tabular}
  \caption{Results of GPT-4o and LLaVA-OneVision-7B on Blink tasks requiring multiple image inputs, where multiple images are concatenated into a single image during inference. Performance changes compared to the zero-shot baseline with multiple image inputs are indicated in parentheses.}

  \label{tab:compare_multi_image_full}
\end{table}

\section{Case Studies} \label{sec: case_study}
\begin{figure*}[!h]
\centering
\includegraphics[width=1.0\linewidth]{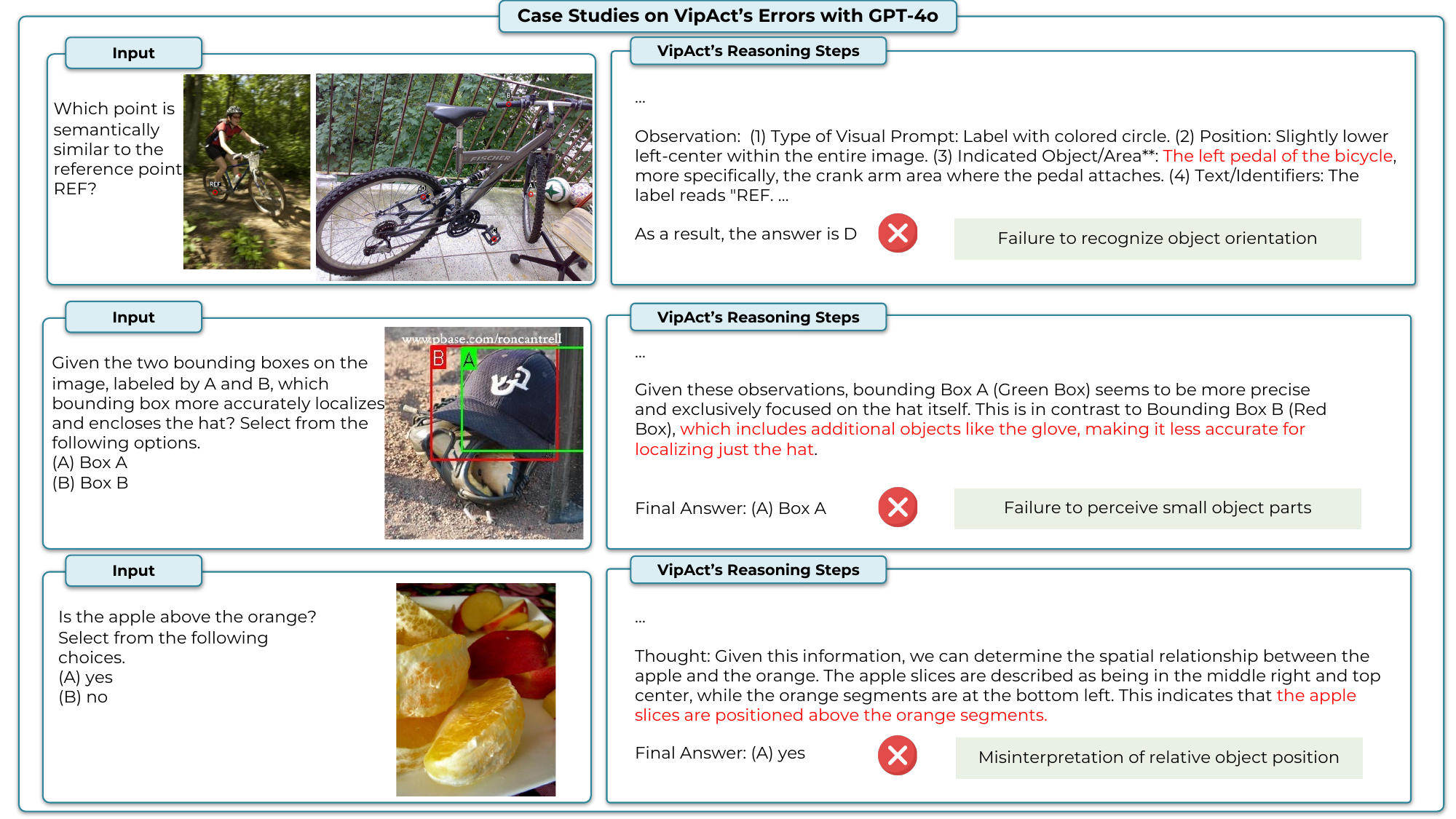}
\caption{Types of error cases in \textsc{VipAct} with their corresponding reasoning steps.}
\label{fig: error_case}
\end{figure*}

\begin{figure*}[h!]
\centering
\includegraphics[width=0.72\linewidth]{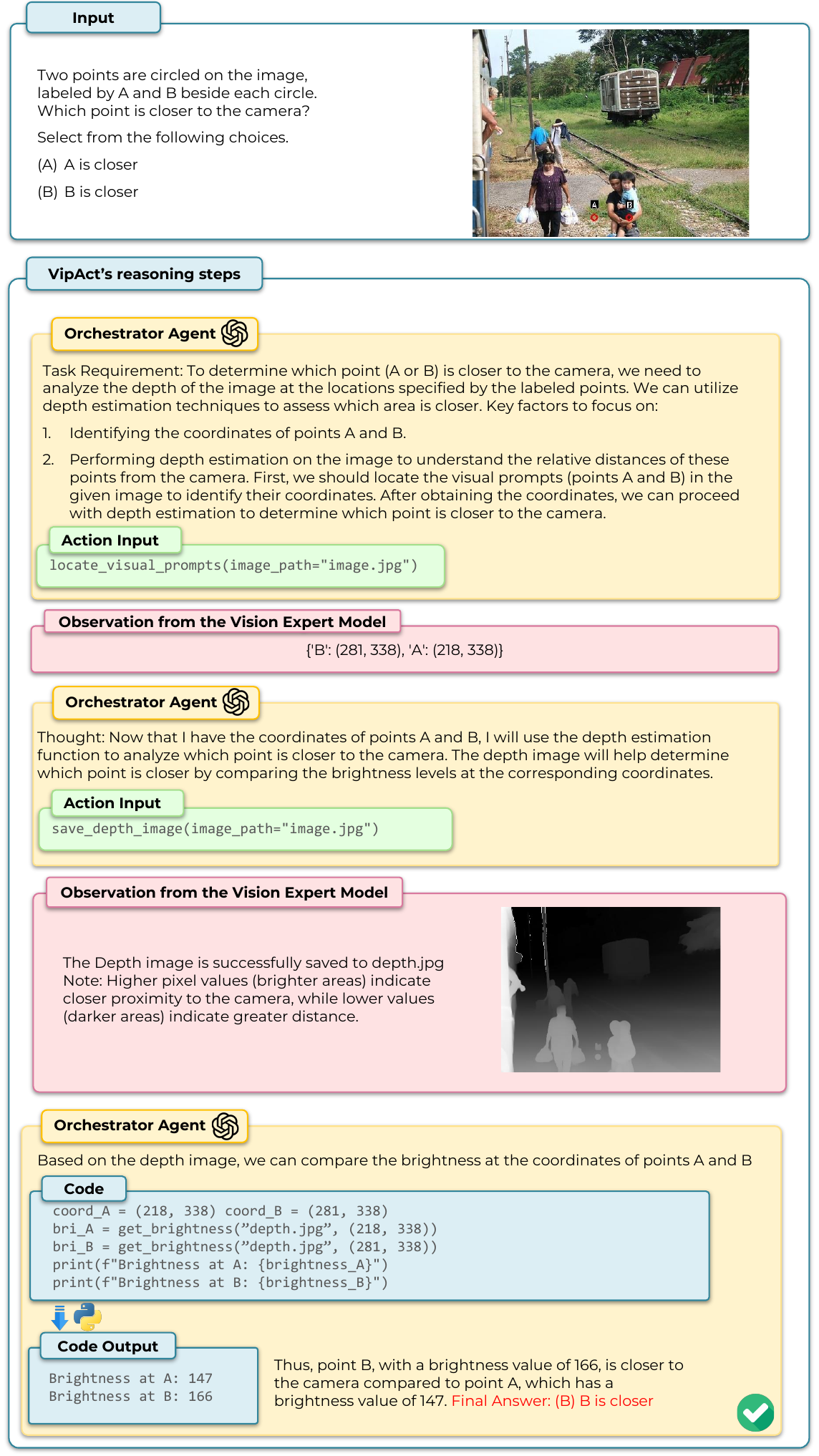}
\caption{Case study of the complete reasoning process in \textsc{VipAct}, incorporating a depth estimation model to assist the VLM in achieving the correct answer.}

\label{fig: case_study_1}
\end{figure*}

\begin{figure*}[h!]
\centering
\includegraphics[width=0.79\linewidth]{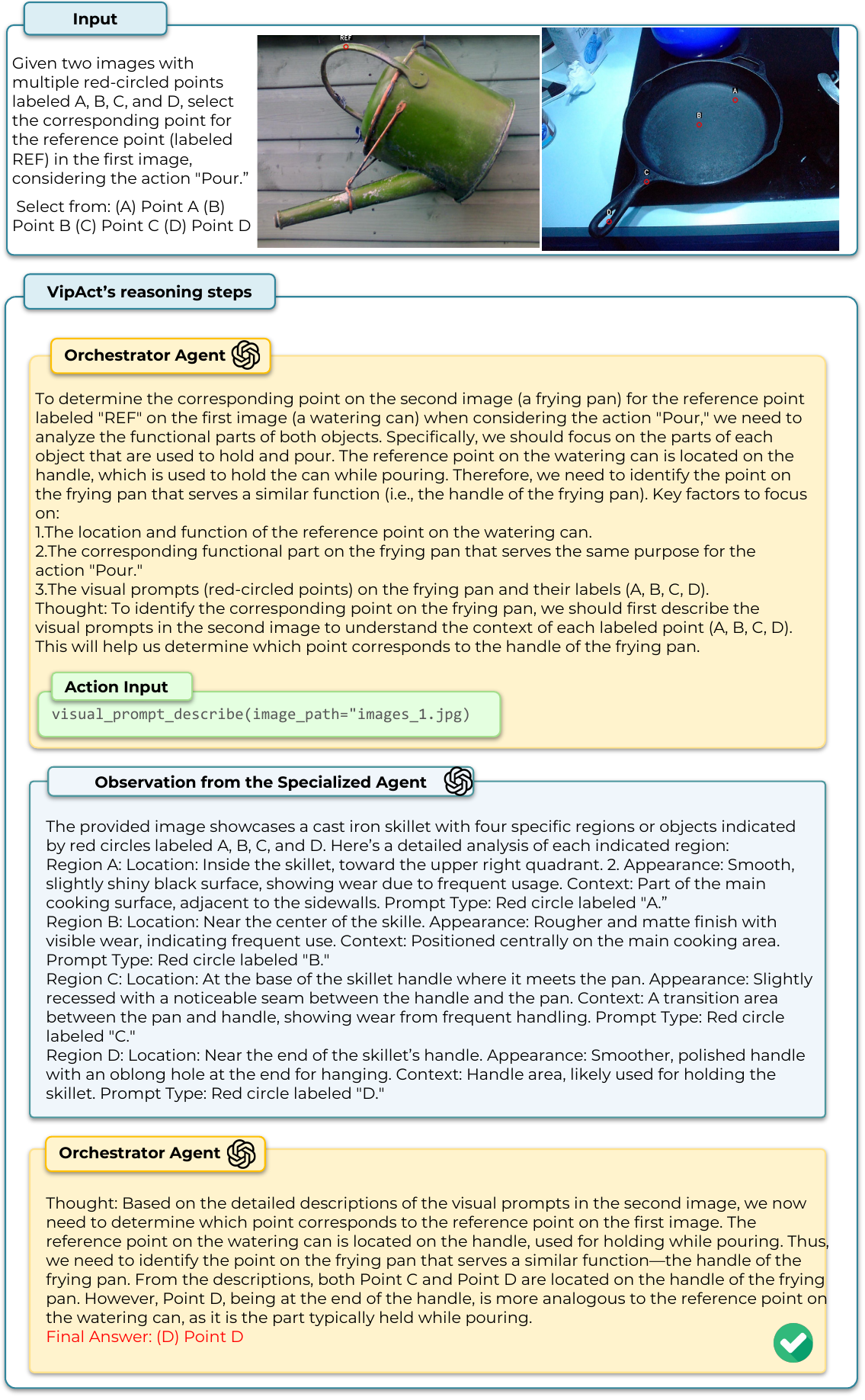}
\caption{Case study of the complete reasoning process in \textsc{VipAct}, incorporating a specialized agent to assist the VLM in achieving the correct answer.}
\label{fig: case_study_2}
\end{figure*}

To intuitively demonstrate the effectiveness of our proposed \textsc{VipAct} and highlight the bottlenecks of current SOTA VLMs, we present a series of case studies showcasing both failure (Figure \ref{fig: error_case}) and success cases (Figures \ref{fig: case_study_1} and \ref{fig: case_study_2}) of our method.

In Figure \ref{fig: error_case}, we observe instances where VLM-based specialized agents in \textsc{VipAct} make reasoning errors, as categorized in Section \ref{sec: error_analysis}. Although \textsc{VipAct} includes an error-handling mechanism to reassess the evidence, these errors can still mislead the orchestrator agent, leading to incorrect inferences. For instance, in the top case of Figure \ref{fig: error_case}, the VLM fails to accurately infer the orientation of the bicycle in the left image, mistakenly identifying the left pedal as the reference point based on the camera's perspective. In the middle case, the VLM overlooks the small portion of the cap's brim, leading to an incorrect prediction. Finally, the bottom case demonstrates how the camera's perspective makes it appear as though the apples are positioned above the orange when in reality, they are on the same plate at the same height. These examples highlight the limitations in visual intelligence exhibited by SOTA VLMs such as GPT-4o, particularly in tasks requiring fine-grained spatial reasoning.

In Figures \ref{fig: case_study_1} and \ref{fig: case_study_2}, we present two examples that demonstrate the complete reasoning process of our \textsc{VipAct}, integrating vision expert models and specialized agents. Figure \ref{fig: case_study_1} illustrates a scenario where the orchestrator agent sequentially invokes vision expert models, including a Visual Prompt Detector and a Depth Estimator, to accurately determine the depth values of two red points in the image, ultimately arriving at the correct answer. In contrast, we observe that GPT-4o is unable to perceive such depth information on its own. Figure \ref{fig: case_study_2} presents a case where no existing vision tools can directly solve the problem. Here, the orchestrator agent introduces a specialized agent specifically designed for visual prompt description. This agent provides a detailed analysis of each visual prompt (marked by red circles) in the second image, leading to the correct prediction. These two examples effectively illustrate the strength of our \textsc{VipAct} framework in integrating vision expert models and specialized agents to enhance reasoning capabilities.

\section{Few-shot In-context Learning}\label{sec: few-shot}
In this section, we examine the effectiveness of few-shot in-context learning in visual perception tasks using various VLMs, including GPT-4o and LLaVA-OneVision-7B. Following previous works \citep{brown2020language, alayrac2022flamingo, awadalla2023openflamingo, zhao2023mmicl, jiang2024many}, we append a series of (image(s), question, answer) triplets—ranging from 1 to 5—before the test query, within the overall instruction. This setup has been shown to enhance performance in LLMs on a wide range of NLP tasks. Additionally, prior research indicates that LLMs can be sensitive to the selection of in-context exemplars \citep{nguyen2023context, zhang-etal-2022-active, agrawal-etal-2023-context, chen2023understanding, zhang2023makes}. To explore this, we employ three different strategies for exemplar selection: (1) Randomly select a specified number of exemplars.
(2) Select exemplars based on top-K similarity using the averaged CLIP embedding of images, which captures both textual semantics and visual information \citep{radford2021learning}.
(3) Select exemplars based on top-K similarity using ViT embeddings \citep{alexey2020image}, which focus purely on visual features. 

Table \ref{tab: few_shot_gpt4o} presents the results of few-shot in-context learning with GPT-4o on the Blink benchmark. We observe that for certain tasks, such as object counting and spatial relations, few-shot learning significantly decreases performance compared to other baselines (see Table \ref{tab: main_result}). However, for tasks like visual correspondence, few-shot in-context learning yields competitive results. Interestingly, as the number of shots increases, no consistent performance trend emerges across the different retrieval methods. Moreover, we do not observe significant or consistent performance differences between the retrieval strategies.

Table \ref{tab: few_shot_llava} shows the results of few-shot in-context learning with LLaVA-OneVision-7B on Blink. Here, we find that performance on almost all sub-tasks is not significantly better than random guessing, even for tasks like object counting and spatial relations, where this model performs much better in baseline settings. Further examination of the outputs reveals that the positional biases identified in Section \ref{sec: other_vlms} persist and even worsen with few-shot prompting, as the model tends to predict the first option in most cases.

In conclusion, while few-shot in-context learning can be effective for some visual perception tasks with GPT-4o, it does not consistently outperform zero-shot baselines and can sometimes negatively impact performance. Additionally, retrieval strategies based on different embedding spaces do not show a clear advantage. For the open-source VLM LLaVA-OneVision-7B, few-shot in-context learning offers no noticeable benefits on these tasks and may even amplify existing biases, further degrading performance.

\section{Exploring the Importance of Multiple Image Inputs to VLMs} \label{sec: multiple_image}

Understanding the relationships between multiple images is crucial for certain visual perception tasks and real-world applications. However, only a few closed-source VLMs \citep{reid2024gemini} and a very limited number of open-source VLMs natively support multiple image inputs. For models that do not support this feature, the common practice is to concatenate multiple images into a single image with added margins and input this combined image into the VLM. To investigate this problem, we conduct experiments using concatenated images for tasks requiring multiple image inputs, utilizing both GPT-4o and LLaVA-OneVision-7B. As shown in Table \ref{tab:compare_multi_image_full}, we observe a noticeable decline in performance for both models when multiple images are concatenated into a single image. This decline is particularly consistent with GPT-4o, indicating that concatenating images introduces challenges that these VLMs struggle to handle effectively. This suggests that native support for multiple image inputs is important for maintaining performance, and concatenating images is not the ideal practice for VLMs.

\begin{table*}[!t]
  \centering
  \begin{tabular}{lcccccccccc}
    \toprule
    \textbf{Method (\# of shots)} & \textbf{Sim} & \textbf{Count} & \textbf{Depth} & \textbf{Jig} & \textbf{Fun.C} & \textbf{Sem.C} & \textbf{Spat} & \textbf{Local} & \textbf{Vis.C} & \textbf{Multi-v} \\
    \midrule
    \rowcolor[rgb]{0.93,0.93,0.93} 
    \multicolumn{11}{l}{\textit{Randomly Choose One From the Options}} \\
    Random \cellcolor{orange!10} & 50.00 & 25.00 & 50.00 & 50.00 & 25.00 & 25.00 & 50.00 & 50.00 & 25.00 & 50.00 \\
    \rowcolor[rgb]{0.93,0.93,0.93} 
    \multicolumn{11}{l}{\textit{Randomly Select In-context Exemplars}} \\
    1-shot \cellcolor{green!10} & 65.93 & 25.00 & 71.77 & 64.00 & 60.77 & 56.12 & 45.45 & 61.48 & 86.05 & 48.12 \\
    2-shot \cellcolor{green!10} & 42.22 & 25.83 & 73.39 & 62.00 & 58.46 & 58.99 & 47.55 & 58.20 & 88.37 & 55.64 \\
    3-shot \cellcolor{green!10}  & 52.59 & 26.67 & 51.61 & 64.00 & 57.69 & 57.55 & 47.55 & 60.66 & 88.37 & 45.11 \\
    4-shot \cellcolor{green!10} & 64.44 & 21.67 & 66.13 & 61.33 & 54.62 & 55.40 & 46.85 & 61.48 & 88.37 & 50.38 \\
    5-shot \cellcolor{green!10}  & 56.30 & 30.00 & 70.16 & 61.33 & 60.77 & 59.71 & 49.65 & 59.84 & 87.79 & 53.38 \\
    \rowcolor[rgb]{0.93,0.93,0.93} 
    \multicolumn{11}{l}{\textit{Select In-context Exemplars Based on CLIP Embedding Similarity}} \\
    1-shot \cellcolor{blue!10} & 78.52 & 20.00 & 66.13 & 52.00 & 56.15 & 56.12 & 44.06 & 58.20 & 87.79& 51.88 \\
    2-shot \cellcolor{blue!10} & 60.00 & 30.00 & 61.29& 60.67 & 54.62 & 53.96 & 47.55 & 63.11 & 84.88 & 51.88 \\
    3-shot \cellcolor{blue!10}  & 52.59 & 26.67 & 59.68& 66.00 & 59.23 & 54.68 & 46.15& 61.48 &89.53 & 51.13 \\
    4-shot \cellcolor{blue!10} &57.04 & 31.67 & 68.55 & 66.00 & 55.38& 56.12 & 45.45 & 63.11 & 88.95 & 56.40 \\
    5-shot \cellcolor{blue!10}  & 60.00 & 25.00 & 64.52 & 62.67 & 58.46 & 53.24 & 47.55 & 59.84 & 87.21 & 54.89 \\
    \rowcolor[rgb]{0.93,0.93,0.93} 
    \multicolumn{11}{l}{\textit{Select In-context Exemplars Based on VIT Embedding Similarity}} \\
    1-shot \cellcolor{cyan!10} & 73.33 & 21.67 & 66.94 & 55.33 & 56.15 & 49.64 & 46.85 & 56.56 & 91.28 & 48.87 \\
    2-shot \cellcolor{cyan!10} & 63.70 & 28.33 & 62.10 & 60.00 & 57.69 & 51.80 & 47.55 & 63.93 & 88.37 & 52.63 \\
    3-shot \cellcolor{cyan!10}  & 57.78 & 27.50 & 62.90 & 64.67 & 57.69 & 53.24 & 46.85 & 60.66 & 89.53 & 50.38 \\
    4-shot \cellcolor{cyan!10} & 46.67 & 30.83 & 61.29 & 64.67 & 56.92 & 53.24 & 48.25 & 59.02 & 89.53 & 48.87 \\
    5-shot \cellcolor{cyan!10}  & 54.07 & 30.00 & 66.13 & 68.00 & 60.77 & 51.08 & 45.45 & 61.40 & 87.79 & 51.13 \\

    \bottomrule
  \end{tabular}
  \caption{Few-shot in-context learning results on the Blink dataset using GPT-4o, evaluated with varying numbers of exemplars and three retrieval methods.
  }
  \label{tab: few_shot_gpt4o}
\end{table*}

\begin{table*}[!t]
  \centering
  \begin{tabular}{lcccccccccc}
    \toprule
    \textbf{Method (\# of shots)} & \textbf{Sim} & \textbf{Count} & \textbf{Depth} & \textbf{Jig} & \textbf{Fun.C} & \textbf{Sem.C} & \textbf{Spat} & \textbf{Local} & \textbf{Vis.C} & \textbf{Multi-v} \\
    \midrule
    \rowcolor[rgb]{0.93,0.93,0.93} 
    \multicolumn{11}{l}{\textit{Randomly Choose One From the Options}} \\
    Random \cellcolor{orange!10} & 50.00 & 25.00 & 50.00 & 50.00 & 25.00 & 25.00 & 50.00 & 50.00 & 25.00 & 50.00 \\
    \rowcolor[rgb]{0.93,0.93,0.93} 
    \multicolumn{11}{l}{\textit{Randomly Select In-context Exemplars}} \\
    1-shot \cellcolor{green!10} & 47.41 & 13.33 & 52.42 & 44.67 & 21.54 & 32.37 & 41.96 & 43.44 & 29.65 & 44.36 \\
    2-shot \cellcolor{green!10} & 47.41 & 2.50 & 54.03 & 52.00 & 22.31 & 32.37 & 38.46 & 43.44 & 29.65 & 55.64 \\
    3-shot \cellcolor{green!10}  & 47.41 & 5.83 & 53.23 & 52.67 & 22.31 & 32.37 & 48.95 & 43.44 & 29.65 & 44.36 \\
    4-shot \cellcolor{green!10} & 47.41 & 3.33 & 52.42 & 52.00 & 22.31 & 32.37 & 45.45 & 43.44 & 29.65 & 44.36 \\
    5-shot \cellcolor{green!10}  & 47.41 & 17.50 & 54.84 & 50.67 & 22.31 & 30.94 & 45.45 & 43.44 & 29.65 & 44.36 \\
    \rowcolor[rgb]{0.93,0.93,0.93} 
    \multicolumn{11}{l}{\textit{Select In-context Exemplars Based on CLIP Embedding Similarity}} \\
    1-shot \cellcolor{blue!10} & 47.41 & 8.33 & 56.45 & 51.33 & 21.54 & 28.06 & 39.16 & 43.44 & 24.42 & 45.11 \\
    2-shot \cellcolor{blue!10} & 47.41 & 8.33 & 54.84 & 51.33 & 22.31 & 25.18 & 39.86 & 43.44 & 27.91 & 30.08 \\
    3-shot \cellcolor{blue!10}  & 47.41 & 10.83 & 53.23 & 50.67 & 20.77 & 26.62 & 39.16 & 43.44 & 27.33 & 28.57 \\
    4-shot \cellcolor{blue!10} & 47.41 & 10.83 & 52.42 & 51.33 & 23.08 & 29.50 & 39.86 & 43.44 & 27.91 & 33.83 \\
    5-shot \cellcolor{blue!10}  & 47.41 & 11.67 & 52.42 & 52.67 & 20.77 & 28.06 & 39.86 & 43.44 & 24.42 & 35.34 \\
    \rowcolor[rgb]{0.93,0.93,0.93} 
    \multicolumn{11}{l}{\textit{Select In-context Exemplars Based on VIT Embedding Similarity}} \\
    1-shot \cellcolor{cyan!10} & 47.41 & 8.33 & 56.45 & 51.33 & 21.54 & 28.06 & 37.06 & 43.44 & 24.42 & 14.29 \\
    2-shot \cellcolor{cyan!10} & 47.41 & 8.33 & 54.84 & 50.67 & 22.31 & 25.18 & 38.46 & 43.44 & 27.91 & 30.08 \\
    3-shot \cellcolor{cyan!10}  & 47.41 & 10.83 & 53.23 & 50.67 & 20.77 & 26.62 & 39.86 & 43.44 & 27.33 & 28.57 \\
    4-shot \cellcolor{cyan!10} & 47.41 & 10.00 & 52.42 & 50.67 & 23.08 & 29.50 & 39.86 & 43.44 & 27.91 & 28.57 \\
    5-shot \cellcolor{cyan!10}  & 47.41 & 10.83 & 52.42 & 52.00 & 20.77 & 28.06 & 41.96 & 43.44 & 24.42 & 34.59 \\

    \bottomrule
  \end{tabular}
  \caption{%
  Few-shot in-context learning results on the Blink dataset using LLaVa-OneVision-7B, evaluated with varying numbers of exemplars and three retrieval methods.
  }
  \label{tab: few_shot_llava}
\end{table*}

\section{Detailed \textsc{VipAct} Algorithm and Function Definitions} \label{sec:func_in_algorithm}

This section presents Algorithm \ref{alg:vipact-framework}, which outlines the complete process of our \textsc{VipAct} framework, providing a clearer illustration of its workflow. To facilitate a comprehensive understanding, Table \ref{tab:vipact_functions} summarizes the detailed explanations of the functions used in the algorithm. Each function plays a crucial role in orchestrating the interactions between the orchestrator agent, specialized agents, and vision expert models within the \textsc{VipAct} framework.

\begin{table*}[!h]
  \centering
  \setlength{\tabcolsep}{3pt} 
  \begin{tabular}{lccccccccccc}
    \toprule
    \textbf{Condition} & \textbf{Sim} & \textbf{Count} & \textbf{Depth} & \textbf{Jig} & \textbf{Fun.C} & \textbf{Sem.C} & \textbf{Spat} & \textbf{Local} & \textbf{Vis.C} & \textbf{Multi-v} & \textbf{Overall} \\
    \midrule
    Image & 81.48 & 70.00 & 90.80 & 68.00 & 61.50 & 60.40 & 86.70 & 63.11 & 91.28 & 62.63 & 73.59 \\
    No image or description input & 77.78 & 59.71 & 69.35 & 61.33 & 53.85 & 51.08 & 83.22 & 60.66 & 78.49 & 48.12 & 64.36 \\
    Description & 79.32 & 62.72 & 73.45 & 62.37 & 54.02 & 52.46 & 83.22 & 61.34 & 81.35 & 51.42 & 66.17 \\
    Image + description & 81.48 & 70.48 & 90.80 & 67.52 & 62.45 & 61.32 & 84.32 & 62.67 & 91.28 & 63.34 & 73.57 \\
    \bottomrule
  \end{tabular}
  \caption{Performance comparison across different input conditions using LLaVa-OneVision-7B on the Blink dataset.}
  \label{tab:additional_visual}
\end{table*}

\begin{table*}[!h]
  \centering
  \setlength{\tabcolsep}{4pt} 
  \begin{tabular}{lcccccccccc}
    \toprule
    \textbf{Method} & \textbf{Sim} & \textbf{Count} & \textbf{Depth} & \textbf{Jig} & \textbf{Fun.C} & \textbf{Sem.C} & \textbf{Spat} & \textbf{Local} & \textbf{Vis.C} & \textbf{Multi-v} \\
    \midrule
    MM-ReAcT (with same tools) & - & 30.00 & 8.69 & - & - & - & 63.64 & 5.41 & - & - \\
    ViperGPT (with same tools) & - & 29.17 & 3.01 & - & - & - & 48.95 & 23.57 & - & - \\
    VisProg (with same tools) & - & 3.33 & 5.31 & - & - & - & 31.47 & 18.92 & - & - \\
    \rowcolor[rgb]{0.93,0.93,0.93}
    \textsc{VipAct} & \textbf{81.48} & \textbf{70.00} & \textbf{90.80} & \textbf{68.00} & \textbf{61.50} & \textbf{60.40} & \textbf{86.70} & \textbf{63.11} & \textbf{91.28} & \textbf{62.63} \\
    \bottomrule
  \end{tabular}
  \caption{Performance comparison of baseline methods integrated with the same vision expert tools as \textsc{VipAct}. \textsc{VipAct} significantly outperforms these baselines due to its cohesive multi-agent design and structured integration of vision expert models.}
  \label{tab:tool_set_fairness}
\end{table*}

\section{In-depth Analysis of Visual Input} \label{sec:visual_input_additional}

To better understand the impact of visual input on the performance of our framework, we conducted an ablation study comparing three input conditions:

\begin{enumerate}
    \item \textbf{Image input only.}
    \item \textbf{Textual description input only}, generated using GPT-4o with the prompt: “Please generate a detailed description of the following image [IMAGE].”
    \item \textbf{Both image and textual description input.}
\end{enumerate}

The results, presented in Table~\ref{tab:additional_visual}, indicate that relying solely on textual descriptions leads to lower performance compared to using image input. Although combining both image and textual descriptions provides slight improvements in some cases, the overall performance gain remains minimal. This suggests that textual descriptions alone may fail to capture critical fine-grained visual details and background elements necessary for accurate task execution.

Further analysis reveals that textual descriptions often emphasize objects while overlooking contextual elements, leading to potential biases and suboptimal decision-making in downstream tasks. In contrast, direct image input allows the orchestrator agent to accurately analyze spatial relationships and identify key visual features required for precise task planning.

These findings emphasize the importance of incorporating direct visual input for robust and contextually grounded reasoning. While textual descriptions can supplement visual input, they are insufficient as standalone inputs for tasks requiring fine-grained perception and reasoning.

\section{Analysis of Tool Set Fairness in Comparisons} \label{sec:tool_set_fairness}

We conducted additional experiments to evaluate the fairness of tool comparisons by integrating our vision expert models into baseline frameworks while preserving their original logic. Unlike many existing visual programming methods, which tightly couple tool usage with internal logic, \textsc{VipAct}’s modular design allows for seamless plug-and-play tool integration with minimal effort, requiring only the definition of a standard Python function header.

Table~\ref{tab:tool_set_fairness} presents the results of these experiments. While incorporating our vision expert tools into baseline frameworks resulted in minor performance improvements in tasks such as depth estimation and object localization, the overall performance of these methods remained significantly lower than \textsc{VipAct}. This disparity arises from the persistent limitations in the underlying logic of these frameworks, which hinder their ability to effectively leverage additional tools.

These results highlight the advantages of \textsc{VipAct}’s cohesive design, which combines multi-agent collaboration and a structured approach to integrating vision expert models. By leveraging these elements, \textsc{VipAct} achieves superior performance and robust generalization across diverse tasks, even when compared to enhanced versions of baseline methods.

\section{Prompt Design} \label{sec: prompt}
In this section, we present the complete prompt designs used in our experiments, including the Initial Prompt for the orchestrator agent and the distinct prompt designs for the three specialized agents described in Section \ref{sec: method}.

\begin{figure*}[t] % or \begin{table*}[t]
  \centering
\begin{tcolorbox}[
  title = {Initial Prompt for Orchestrator Agent},
  colback = Apricot!25!white,
  colframe = BrickRed!75!black,
  breakable,
  enhanced
]
You are a helpful AI agent and please answer the following question based on the image. You have access to the following tools:

\{tools\}

Additionally, if you want to use python code, you can use the following functions:

\begin{verbatim}
def image_comparison(image_paths: list, focus: str = None):
        '''
        Compares multiple images and generates a detailed 
        analysis of their similarities and differences,
        with an optional focus on specific objects, elements, 
        or aspects.

        Parameters
        ----------
        image_paths : list
            A list of file paths for the input images to 
            be compared.
        focus : str, optional
            The specific objects, elements, or aspects that 
            the comparison should focus on.
            If None, a general comparison is generated.

        Example
        --------
            >>> image_comparison(image_paths=["image1.jpg", 
            "image2.jpg"], focus="the cars")
        '''
\end{verbatim}
\end{tcolorbox}
\end{figure*}

\newpage

\begin{figure*}[t] % or \begin{table*}[t]
  \centering
\begin{tcolorbox}[
  title = {Initial Prompt for Orchestrator Agent (Cont'd)},
  colback = Apricot!25!white,
  colframe = BrickRed!75!black,
  breakable,
  enhanced
]
\small
\begin{verbatim}
def image_captioning(image_path: str, focus: str = None):
        '''
        Generates a detailed caption for the provided image, 
        with an optional focus on specific objects, elements or 
        other perspectives that are directly related to solving 
        the problem.

        Parameters
        ----------
        image_path : str
            The file path of the input image.
        focus : str, optional
            The specific objects or elements that the caption 
            should focus on. If None, a general caption is 
            generated.

        Example
        --------
            >>> image_captioning(image_path="image.jpg")
            >>> image_captioning(image_path="image.jpg", 
            focus="a red car and the background buildings")
        '''

    def visual_prompt_describe(image_path: str = "image.jpg"):
        '''
        Analyzes the provided image and describes the specific 
        locations and characteristics of various visual prompts

        This function uses a language model to generate a 
        detailed description of visual prompts present in the 
        image, such as colored circles, bounding boxes, arrows, 
        highlights, or textual labels.

        Parameters
        ----------
        image_path : str
            The file path of the input image.
        
        Example
        --------
            >>> visual_prompt_describe(image_path="image.jpg") 
\end{verbatim}
\end{tcolorbox}
\end{figure*}
\newpage

\begin{figure*}[t] % or \begin{table*}[t]
  \centering
\begin{tcolorbox}[
  title = {Initial Prompt for Orchestrator Agent (Cont'd)},
  colback = Apricot!25!white,
  colframe = BrickRed!75!black,
  breakable,
  enhanced
]
\small
\begin{verbatim}
def save_depth_image(image_path: str = "image.jpg", 
    saved_path: str = "depth.jpg"):
        '''
        Estimates the depth of an input image, saves the 
        resulting depth image to a specified path, 
        and prints out the saved path in a structured format.
        
        Note: In the processed depth estimation image, brighter 
        areas represent objects closer to the camera,
        while darker areas represent objects farther from the 
        camera. For pixel values, higher values (brighter areas)
        indicate closer proximity to the camera, while lower 
        values (darker areas) indicate greater distance.

        Parameters
        ----------
        image_path : str, optional
            The file path of the input image. 
            
        saved_path : str, optional
            The file path where the resulting depth image will 
            be saved. You should make sure the saved image is 
            in the same directory as the input image.

        Example 
        --------
            >>>  save_depth_image(image_path = "image.jpg", 
            saved_path = "depth.jpg")
        '''
    
    def locate_visual_prompts(image_path: str = "image.jpg"):
        '''
        Analyzes the provided image to identify and accurately 
        locate two specific regions labeled 'A' and 'B'.
        This function detects the visual prompts of red circles 
        and print out their coordinates.

        Parameters
        ----------
        image_path : str
            The file path of the input image to be processed.

        Example
        -------
            >>> locate_visual_prompts("images/image.jpg")
        '''
\end{verbatim}
\end{tcolorbox}
\end{figure*}

\newpage
\begin{figure*}[t] % or \begin{table*}[t]
  \centering

\begin{tcolorbox}[
  title = {Initial Prompt for Orchestrator Agent (Cont'd)},
  colback = Apricot!25!white,
  colframe = BrickRed!75!black,
  breakable,
  enhanced
]
\small
\begin{verbatim}
 def compute_clip_similarity(image_path1: str, 
    image_path2: str) -> float:
        '''
        Computes the cosine similarity between the CLIP 
        embeddings of two images.

        Parameters
        ----------
        image_path1 : str
            The file path of the first input image.
        image_path2 : str
            The file path of the second input image.

        Returns
        -------
        float
            The cosine similarity score between the two image 
            embeddings (-1 to 1).

        Example
        -------
            >>> similarity = 
            compute_clip_similarity("image1.jpg", "image2.jpg")
        '''

    def segment_image(image_path: str, save_path: str = None) 
    -> str:
        '''
        Segments the input image using the SAM model and  
        returns the path to the processed image.

        Parameters
        ----------
        image_path : str
            The file path of the input image to be segmented.
        save_path : str, optional
            The file path where the segmented image will be 
            saved. If None, a default path is used.

        Returns
        -------
        str
            The file path of the saved segmented image.

        Example
        -------
            >>> segmented_img_path = segment_image("input.jpg", 
            "segmented.jpg")
\end{verbatim}
\end{tcolorbox}
\end{figure*}

\begin{figure*}[t] % or \begin{table*}[t]
  \centering
\begin{tcolorbox}[
  title = {Initial Prompt for Orchestrator Agent (Cont'd)},
  colback = Apricot!25!white,
  colframe = BrickRed!75!black,
  breakable,
  enhanced
]
All function implementations are available in the execution environment and you can just call the function without the need to define it. \\
MUST strictly use the following format:

Question: [The input question you must answer]\\
Image: [The path of the image, which you may use in external tools]\\
Task Requirement: [You should provide a comprehensive analysis of the criteria to choose between each option. Include key factors to focus on in solving this task, such as specific visual elements, data points, trends, patterns, and any contextual information that might influence the decision. You can also try to decompose the problem into several key subproblems, with clues inferred from the following steps.]\\
Thought: [Your reasoning about the question or the last iteration's observations. You should prioritize to think about which tools to use (and which parameters to input) and if you believe no existing tools will help further, use your own knowledge to reason towards the final answer. If there is no observation from the last iteration's tool calling, you should examine the format of tool calling and recall the tool with proper format]\\
Action Input: [MUST be some of the functions above within a Python block with nothing else. You should figure out which function to use and what are the input parameters.]\\
Observation: [The output of the called function.]\\
... (Repeat Thought/Action/Action Input/Observation as needed, you may need to call the tools multiple times if there are multiple images in the input)
Thought: [Your final reasoning based on all information gathered]\\
Final Answer: [You MUST provide a clear answer from the above options without any ambiguity. If a perfect answer is not available, you MUST select the closest possible option.]

Begin! Let's work on the following question! Please remember NOT to estimate any coordinates in the image within the code.\\
Question: \textcolor{blue}{\{question\}} \\
Image: \textcolor{blue}{\{image\}}
Task Requirement: (you should start to generate this to begin the iterations)

\end{tcolorbox}
\end{figure*}

\begin{figure*}[t] % or \begin{table*}[t]
  \centering
\begin{tcolorbox}[
  title = {Prompt for Focused Image Captioning Agent},
  colback = Apricot!25!white,
  colframe = BrickRed!75!black,
  breakable,
  enhanced
]
\small
Please analyze the provided image and generate a comprehensive, detailed caption that focuses specifically on "\textcolor{blue}{\{focus\}}". Your caption should:

    1. Identify and describe the specified focus objects or elements in the image, including: 
    \begin{itemize}
        \item Quantity (the total number of such object)
        \item  Appearance (color, size, shape, texture)
        \item  Position within the image
        \item  Relation to other objects (if applicable)
    \end{itemize}
    2. For the focus objects or elements, describe any actions or events taking place, involving any of them. \\
    3. Mention the overall setting or background of the image, especially in relation to the focus.\\
    4. Include relevant details about lighting, shadows, and any visible textures.\\
    5. If there are people or animals in the focus area, describe their appearances, poses, and any visible expressions.

    Your goal is to create an extremely detailed and thorough caption that gives a complete understanding of the image's content with an emphasis on the specified focus, as if you're describing it to someone who cannot see it. Don't leave out any visible elements related to the focus, no matter how minor they might seem.

Image: \textcolor{blue}{\{image\}}
\end{tcolorbox}

\end{figure*}

\newpage

\begin{figure*}[t] % or \begin{table*}[t]
  \centering
\begin{tcolorbox}[
  title = {Prompt for Focused Image Comparison Agent},
  colback = Apricot!25!white,
  colframe = BrickRed!75!black,
  breakable,
  enhanced
]
\small
Please analyze the provided images and generate a comprehensive, detailed comparison that focuses specifically on "\textcolor{blue}{\{focus\}}". Your comparison should:

    1. Identify and describe the specified focus ({focus}) in all images, including:

    \begin{itemize}
        \item Presence or absence in each image (if applicable)
        \item Quantity (if applicable)
        \item Position within each image
        \item Relation to other objects (if applicable)
    \end{itemize}
    2. Compare the overall setting or background of the images, but only as it relates to the focus. \\
    2. Summarize the similarities and differences of the focus elements across all images. \\
    3. Describe any changes in actions, events, or states related to the focus elements (if applicable). \\
    5. Analyze differences in lighting, shadows, and visible textures that affect the focus elements.

    Your goal is to create a detailed and thorough comparison that gives a complete understanding of how the specified focus elements differ or remain similar across all the provided images. Concentrate primarily on the focus area and only mention other elements if they directly relate to or impact the focus.

    Organize your comparison in a clear, structured manner, addressing the focus area in each image in turn and then providing an overall summary of the similarities and differences.

Image: \textcolor{blue}{\{image\}}
\end{tcolorbox}
\end{figure*}

\begin{figure*}[t] % or \begin{table*}[t]
  \centering
  
\begin{tcolorbox}[
  title = {Prompt for Visual Prompt Description Agent},
  colback = Apricot!25!white,
  colframe = BrickRed!75!black,
  breakable,
  enhanced
]

Please analyze the provided image, emphasizing the specific regions or objects indicated by visual prompts such as colored circles, bounding boxes, arrows, highlights, or textual labels. The most critical aspect of your analysis should be a detailed description of these indicated areas. For each visual prompt:

1. Most importantly, provide an extremely detailed description of the exact region or object being indicated. This is the primary focus of your analysis. Include:
\begin{itemize}
    \item Precise location within the larger object or scene
    \item Comprehensive details about its appearance (color, texture, shape, size)
    \item Any unique features or characteristics
    \item Its context and relationship to surrounding elements
\end{itemize}

2. The type of visual prompt used (e.g., circle, box, arrow, highlight, label).\\
3. The position of the prompt within the entire image (e.g., top left, center, bottom right).\\
4. Any text or identifiers associated with the prompt (e.g., labels like 'A', 'B', numbers, or short descriptions).
Remember, the most crucial part of your response should be the in-depth description of the specific region or object highlighted by each prompt. Provide enough detail that someone could understand exactly what part of the image is being emphasized without seeing the visual prompt itself.\\
Ensure your description of these indicated regions is as comprehensive as possible, covering every relevant visual aspect. Your goal is to provide a thorough understanding of the highlighted areas, allowing others to easily grasp the significance of each visual prompt in the image.\\
Image: \textcolor{blue}{\{image\}}
\end{tcolorbox}
\end{figure*}

\begin{figure*}[t] % or \begin{table*}[t]
  \centering
  
\begin{tcolorbox}[
  title = {Prompt for Few-shot In-context Learning},
  colback = Apricot!25!white,
  colframe = BrickRed!75!black,
  breakable,
  enhanced
]
\textcolor{blue}{\{The general instruction for the task\}}\\
Here are some examples: \\
Images: \textcolor{blue}{\{example\_images\}} \\
Question: \textcolor{blue}{\{example\_question\}} \\
Answer: \textcolor{blue}{\{example\_answer\}}\\
...\\
Let's try another case!\\
Images: \textcolor{blue}{\{images\}}\\
Question: \textcolor{blue}{\{question\}}\\
Answer: 
\end{tcolorbox}
\end{figure*}

\begin{table*}[htb]
    \centering
    \begin{tabular}{p{4.5cm} p{10.5cm}}
        \toprule
        \textbf{Function} & \textbf{Description} \\ \midrule
        \textsc{FormatPrompt}($\mathcal{V}, q$) & 
        Combines the visual inputs $\mathcal{V}$ and the query $q$ into a structured prompt suitable for the vision-language model $\mathcal{M}$. This ensures that the orchestrator agent receives a well-organized task description for reasoning. \\ \midrule
        
        \textsc{IsTerminated}($\mathcal{S}$) & 
        Checks whether the termination condition has been met based on the current state $\mathcal{S}$. This involves checking for a termination indicator (e.g., \texttt{Final Answer:}) or determining if the maximum number of iterations $K$ has been reached. \\ \midrule
        
        \textsc{IsRequired}($T_i, \mathcal{S}$) & 
        Determines if a specific tool $T_i$ (either a specialized agent or vision expert model) is necessary given the current state $\mathcal{S}$. This involves checking whether tool-use indicators (e.g., \texttt{Action:} or \texttt{Action Input:}) have been generated, guiding the orchestrator agent on whether external tools need to be invoked. \\ \midrule
        
        \textsc{Utility}($T_i, \mathcal{S}$) & 
        Implicitly evaluates the utility of tool $T_i$ in the current context defined by state $\mathcal{S}$. This process involves the orchestrator agent select the most beneficial tool for the next action, based on prior evidence and reasoning steps. \\ \midrule
        
        \textsc{Execute}($T^*, \mathcal{S}$) & 
        Executes the selected tool $T^*$ using the current state $\mathcal{S}$ (arguments extracted from VLM's output at this step) as input. The tool processes the input and returns relevant information, such as image data or analytical results, which are then integrated into the reasoning process. \\ \midrule
        
        \textsc{ContainsVisualData}($\mathcal{O}_t$) & 
        Checks whether the output $\mathcal{O}_t$ from the executed tool includes visual data (e.g., new images or annotations). If visual data is present, it is further processed and incorporated into the reasoning workflow. \\ \midrule
        
        \textsc{ProcessVisualData}($\mathcal{O}_t$) & 
        Processes new visual data from the tool’s output $\mathcal{O}_t$ and integrates it into the existing set of visual inputs $\mathcal{V}$. This involves updating the prompt with new image paths to ensure that the visual data is available for subsequent analysis and reasoning. \\ \midrule
        
        \textsc{InterpretOutput}($\mathcal{R}_t$) & 
        Interprets the output $\mathcal{R}_t$ generated by the VLM $\mathcal{M}$. This step involves converting the raw output into a structured format through rule-based string manipulation, enabling the orchestrator agent to update the task state and inform the next steps. \\ \midrule
        
        \textsc{UpdatePrompt}($\mathcal{P}_{t-1}, \mathcal{O}_t$) & 
        Updates the current prompt $\mathcal{P}_{t-1}$ with new information derived from the tool output $\mathcal{O}_t$. The updated prompt ensures that the next iteration of the VLM has access to the most recent and relevant context, presented in an organized format for accurate reasoning in the next iteration. \\ \midrule
        
        \textsc{UpdateState}($\mathcal{S}, \mathcal{O}_t$) & 
        Updates the current state $\mathcal{S}$ by incorporating new observations and data from the tool or VLM output $\mathcal{O}_t$. This continuous state update allows the system to track progress and adjust its strategy dynamically. \\ \midrule
        
        \textsc{ExtractAnswer}($\mathcal{S}$) & 
        Extracts the final answer $a$ from the final output of the VLM. This step uses rule-based string matching to retrieve the final prediction from the agent's workflow. \\ \bottomrule
    \end{tabular}
    \caption{Function Definitions in Algorithm \ref{alg:vipact-framework}}
    \label{tab:vipact_functions}
\end{table*}

\end{document}